\documentclass[sigconf]{acmart}

\usepackage{booktabs} % For formal tables
\usepackage[english]{babel}
\usepackage[utf8]{inputenc}
\usepackage{amsmath}
\usepackage{amssymb}
\usepackage{mathrsfs}
\usepackage{graphicx}
\usepackage[colorinlistoftodos]{todonotes}
\usepackage{geometry}
\usepackage{hyperref}

\usepackage{graphicx}
\usepackage{subcaption}
\usepackage{natbib}

\usepackage{algorithm}
\usepackage[noend]{algpseudocode}
\usepackage{hyperref}

\usepackage{mathtools}
\usepackage{caption}
\usepackage{dirtytalk}
\usepackage{textcomp, libertine}

\usepackage{balance}

\setcopyright{rightsretained}
\begin{document}
\title{Customer Lifetime Value Prediction Using Embeddings}

% \titlenote{Produces the permission block, and
%   copyright information}
% \subtitle{Extended Abstract}
% \subtitlenote{The full version of the author's guide is available as
%   \texttt{acmart.pdf} document}

\author{Benjamin Paul Chamberlain}
%\authornote{}
\orcid{}
\affiliation{
  \department{Department of Computing}
  \institution{Imperial College London}
}
\email{b.chamberlain14@imperial.ac.uk}

\author{\^{A}ngelo Cardoso}
%\authornote{}
\orcid{}
\affiliation{
  \institution{ASOS.com}
  \city{London}
  \country{United Kingdom}
}

\author{C.H. Bryan Liu}
%\authornote{}
\orcid{}
\affiliation{
  \institution{ASOS.com}
  \city{London}
  \country{United Kingdom}
}

\author{Roberto Pagliari}
%\authornote{}
\orcid{}
\affiliation{
  \institution{ASOS.com}
  \city{London}
  \country{United Kingdom}
}

\author{Marc Peter Deisenroth}
%\authornote{}
\orcid{}
\affiliation{
  \department{Department of Computing}
  \institution{Imperial College London}
}
\email{m.deisenroth@imperial.ac.uk}

% The default list of authors is too long for headers}
\renewcommand{\shortauthors}{B.P. Chamberlain et al.}

\begin{abstract}
% \todo[inline, color=yellow]{To be submitted to the deployed applied data science track at KDD17. The brief is: 
% papers \textbf{describing} designs and implementations of solutions and systems for practical tasks. The primary emphasis is on papers that advance the understanding of, and \textbf{show how to deal with, practical issues related to deploying data science technologies}.
% Deployed - Must describe implementation of a system that solves a significant real-world problem. The focus should be on describing the problem, its significance, decisions and tradeoffs made when making design choices for the solution, deployment challenges, and lessons learned.}
% specify which track this is for
We describe the Customer LifeTime Value (CLTV) prediction system deployed at ASOS.com, a global online fashion retailer.  
% why this matters
CLTV prediction is an important problem in e-commerce where an accurate estimate of future value allows retailers to effectively allocate marketing spend, identify and nurture high value customers and mitigate exposure to losses.
% brief system description
The system at ASOS provides daily estimates of the future value of every customer and is one of the cornerstones of the personalised shopping experience.
% what is the state of the art
The state of the art in this domain uses large numbers of handcrafted features and ensemble regressors to forecast value, predict churn and evaluate customer loyalty. 
% why this is bad
Recently, domains including language, vision and speech have shown dramatic advances by replacing handcrafted features with features that are learned automatically from data.
% what we do
We detail the system deployed at ASOS and show that learning feature representations is a promising extension to the state of the art in CLTV modelling. We propose a novel way to generate embeddings of customers, which addresses the issue of the ever changing product catalogue and obtain a significant improvement over an exhaustive set of handcrafted features.
 
% why we need embeddings
% Good results are only possible by leveraging the large, rich and heterogenous data that can be collected through e-commerce sites. The richest sources of data are the browsing logs from the web and mobile apps. The logs detail any interactions with pages, products or content. In their raw form however, they are difficult to incorporate directly into machine learning systems as they are large, sparse and discrete.
% ephemeral embeddings
% We solve this problem by using the browsing logs to generate an ephemeral embedding for each customer. The embeddings are ephemeral because they use products and customers that can exist on the site for only short periods of time. 
% % solving the time problem
% Unlike the eternal embeddings, used in natural language processing, prediction problems with ephemeral embeddings must handle the emergence of large numbers of new tokens in the prediction period. This is challenging because the dimensions of an embedding vector are exchangeable and so there is no guarantee that parameters learnt in a training period will correspond to the dimensions of the embeddings in the prediction period. 
% \todo[inline]{still need summary line}
% We show how to overcome this problem and  

% what do people currently do 
% what is innovative about our method
% any compelling results
\end{abstract}

%
% The code below should be generated by the tool at
% http://dl.acm.org/ccs.cfm
% Please copy and paste the code instead of the example below. 
%

\copyrightyear{2017}
\acmYear{2017}
\setcopyright{acmlicensed}    
\acmConference{KDD'17}{}{August 13--17, 2017, Halifax, NS, Canada.} 
\acmPrice{15.00}
\acmDOI{http://dx.doi.org/10.1145/3097983.3098123}
\acmISBN{978-1-4503-4887-4/17/08}


% We no longer use \terms command
%\terms{Theory}


\maketitle

\section{Introduction}
% \todo[inline, color=green]{marc: I'm missing in the introduction information about a) What the problem is with existing approaches, and what the key insight is that leads to an improvement. There are some details further down the line, but this needs to come much earlier and punchier.}

% introduction to ASOS
ASOS is a global e-commerce company, based in the UK and specialising in fashion and beauty. The business is entirely online, and products are sold through eight country-specific websites and mobile apps. At the time of writing, there were 12.5 million active customers and the product catalogue contained more than 85,000 items. Products are shipped to 240 countries and territories and the annual revenue for 2016 was £1.4B, making ASOS one of Europe's largest pure play online retailers. 

% the free returns business model
An integral element of the business model at ASOS is free delivery and returns. Free shipping is vital in online clothing retail because customers need to try on items without being charged. Since ASOS do not recoup delivery costs for returned items, customers can easily have negative lifetime value. For this reason the Customer LifeTime Value (CLTV) problem is particularly important in online clothing retail.

% defining CLTV and churn 
Our CTLV system addresses two tightly coupled problems: CLTV and churn prediction. We define a customer as churned if they have not placed an order in the past year. We define CLTV as the sales, net of returns, of a customer over a one year period.  
% system objectives
The objective is to improve three key business metrics: (1) the average customer shopping frequency, (2) the average order size, (3) the customer churn rate. The model supports the first two objectives by allowing ASOS to rapidly identify and nurture high-value customers, who will go on to have high frequency, high-order size, or both. The third objective is achieved by identifying customers at high risk of churn and controlling the amount spent on retention activities.

% describe the state of the art
State-of-the-art CLTV systems use large numbers of handcrafted features and ensemble classifiers (typical random forest regressors), which have been shown to perform well in highly stochastic problems of this kind~\cite{Vanderveld2016, Hardie2006}. However, handcrafted features introduce a human bottleneck, can be difficult to maintain and often fail to utilise the full richness of the data. 
% adding learned features
We show how automatically learned features can be combined with handcrafted features to produce a model that is both aware of  domain knowledge  and can learn rich patterns of customer behaviour from raw data.

\begin{figure}[tb]
\includegraphics[width=0.48\textwidth]{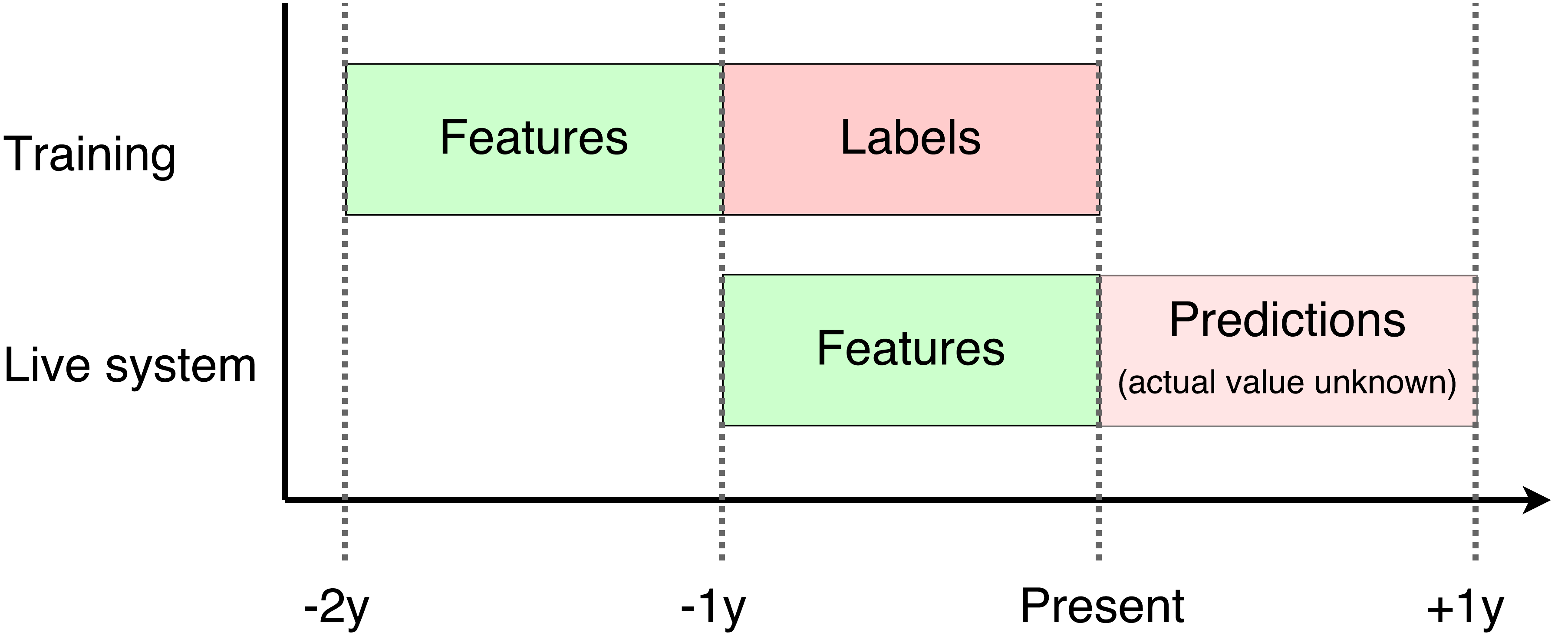}
\caption{Training and prediction time-scales for CLTV. The model is retrained every day using customer data from the past two years. Labels are the net customer spend over the previous year. Model parameters are learned in the training period and used to predict CLTV from new features in the live system.}
\label{fig:cltv_training}
\end{figure}

% outline of the model
The deployed ASOS CLTV system uses the state-of-the-art architecture, which is a  Random Forest (RF) regression model with 132 handcrafted features. To train the forest, labels and features are taken from disjoint periods of time as shown in the top row of Figure~\ref{fig:cltv_training}. The labels are the net spend (sales minus returns) of each customer over the past year. The training labels and features are used to learn the parameters of the RF. The second row of Figure~\ref{fig:cltv_training} shows the live system where the RF parameters from the training period are applied to new features generated from the last year's data to produce prediction for net customer spend over the next year.
% training procedure

\begin{figure}
\includegraphics[width=0.48\textwidth]{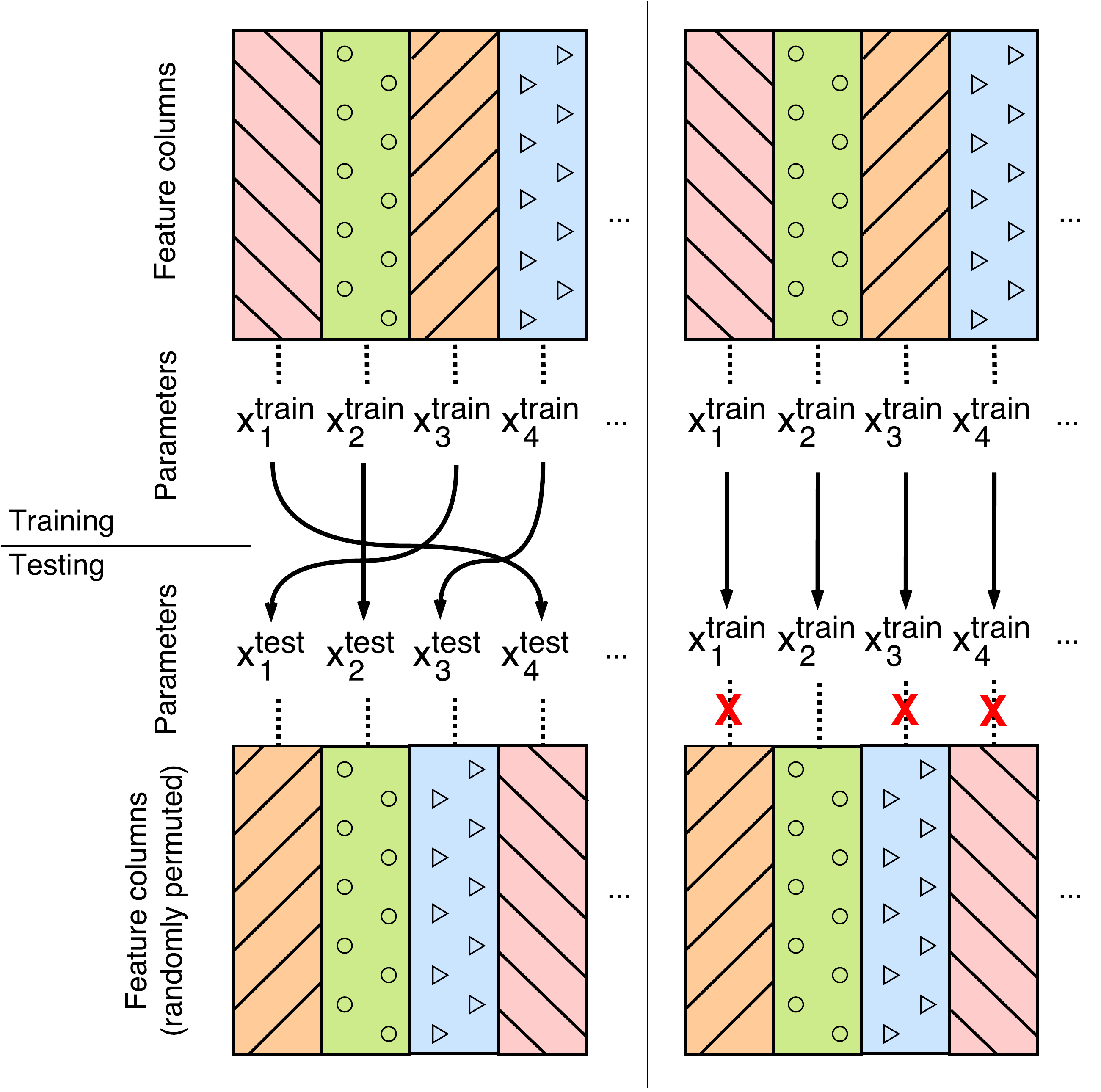}
\caption{Illustration of the challenges of using the components of embedded customer vectors as features for forecasting.  Each column represents a component of the vector representation of customers. We have labelled the columns with the parameters that are learnt at training time. (Left) The vector components randomly permute between train and testing time and hence require different learned parameters. (Right) Applying the learned parameters from training time directly to the embeddings in test time will not work as they are no longer attached to the correct component of the embedded vectors.}
\label{fig:embedding_feature_permutation}
\end{figure}

% We have exhausted the improvements we can get using the standard approach to this problem by exhaustively generating handcrafted features.
We provide a detailed explanation of the challenges and lessons learned in deploying the ASOS CLTV system and describe our latest efforts to improve this system by augmenting it with learned features.
Our approach is inspired by the recent successes of representation learning in the domains of vision, speech and language. We experimented with learning representations directly from data using two different approaches: (1) by training a feedforward neural network on the handcrafted features in a supervised setting (see Section~\ref{sec:embhandcraftfeat}); (2) By augmenting the RF feature set with unsupervised customer embeddings learnt from web and app browsing data (see Section~\ref{sec:embcustsessions}).
% customer embeddings
The novel customer embeddings are shown to  improve CLTV prediction performance significantly compared with our benchmark. Incorporating embeddings into long-term prediction models is challenging because, unlike handcrafted features, the features are not easily identifiable. Figure~\ref{fig:embedding_feature_permutation} illustrates this point using a four dimensional embedding. Each column in the figure corresponds to a different dimension of the embeddings space and also a feature of the random forest model. In the training period we learn parameters for each feature. However, as the features are not labelled and their order randomly permutes from training to test time, it is not possible to map the training parameters to the test features. We describe how this problem can be solved within the neural embedding framework using a form of warm start for the test period embeddings.

Therefore, our main contributions are:
\begin{enumerate}
\item A detailed description of the large-scale CLTV system deployed at ASOS, including a discussion of the architecture, deployment challenges and lessons learned.
\item Analysis of two candidate architectures for hybrid systems that incorporate both handcrafted and learned features
\item Introduction of customer level embeddings and demonstration that these produce a significantly better performance than a benchmark classifier
\item We show how to use neural embeddings for long-term prediction tasks
\end{enumerate}

% \todo[inline]{move this somewhere else}
% % why this is hard
% A naive random forest regressor fails for this task. This is attributable to the unusual distribution of observed CLTVs. A large fraction of customers churn, with a CLTV of zero. The remaining exhibit a power law distributed CLTV spanning six orders of magnitude. 
% \todo[inline]{Fig: quantile function of CLTV at ASOS}
% % what we do that is novel
% We experimented with a number of non-linear regressors including several deep neural architectures \todo[inline]{link to this description}. Our solution was to first learn a regressor for percentiles and then learn a separate mapping from percentiles to monetary values to increase the robustness to outliers, resulting in a lower RMSE than by forecast the monetary value directly.

% outline of the implementation

% what is the novelty

% Any results

\section{Related Work}

Statistical models of customer purchasing behaviour have been studied for decades. Early models were hampered by a lack of data and were often restricted to fitting simple parametric statistical models, such as the Negative Binomial Distribution in the NBD model~\cite{Journal2016}. It was only at the turn of the century, with the advent of large-scale e-commerce platforms that this problem began to attract the attention of machine learning researchers.

\subsection{Distribution Fitting Approaches}

The first statistical models of CLTV were known as ``Buy 'Til You Die'' (BTYD) models. BTYD models place separate parametric distributions on the customer lifetime and the purchase frequency and only require customer recency and frequency as inputs. One of the first  models was the Pareto/NBD model~\cite{Schmittlein1987}, which assumes an exponentially distributed active duration and a Poisson distributed purchase frequency for each customer.
% A branch of CLTV modeling attempts to model the probability of a customer being active and expected number of purchase at a time in future, given their past purchase frequency and time of last purchase in an observation period. The models assume each customer will be ``alive'' (active) for a certain period, and make a certain number of purchase(s) while being ``alive'', both of which specified by some probability distributions respectively. The distribution hyperparameters are estimated with maximum likelihood or moment fitting using past data.
%with rates specified by two independent gamma priors. 
This yields a Pareto-II customer lifetime and negative-binomial (NBD) distributed purchase frequency. Fader et al.~\cite{Fader2005b} replaced the Pareto-II with a Beta-geometric distribution for easier implementation, assuming each customer has a certain probability to become inactive after every purchase.
%(geometric) specified by a beta prior. 
Bemmaor et al.~\cite{Bemmaor2012} proposed the Gamma/Gompertz distribution to model the customer active duration, which is more flexible than the Pareto-II as it can have a non-zero mode and be skewed in both directions.

% RFM models
Recency-Frequency-Monetary Value (RFM) models expand on BTYD by including an additional feature. 
Fader et al.~\cite{Fader2005a} linked the RFM model, which captures the time of last purchase (recency), number of purchases (frequency) and purchase values (monetary value) of each customer to CLTV estimation. Their model assumes a Pareto/NBD model for recency and frequency with purchase values following an independent gamma/gamma distribution. 
% In the gamma/gamma distribution each value is gamma distributed around the associated customer's mean purchase value, which follows another gamma distribution across all customers. The predicted CLTV is then calculated by multiplying the expected number of purchases from the Pareto/NBD model by the expected purchase value from the gamma/gamma model. The hyperparameters are estimated using maximum likelihood from past data.

While successful for the problems that they were applied to, it is difficult to incorporate the vast majority of the customer data available to modern e-commerce companies into the RFM/BYTD framework. It is particularly difficult to incorporate automatically learned or highly sparse features. This motivates the need for machine learning approaches to the problem.

% The models above also assume customers who made the same number of purchases, had last purchase at the same time, and potentially spent the same amount on ASOS are the same; whereas they could possess very different characteristics.

\subsection{Machine Learning Methods}

The most related work to ours is that of~\citet{Vanderveld2016}, which is the first work to explicitly include customer engagement features in a CLTV prediction model. They also address the challenges of learning the complex, non-linear CLTV distribution by solving several simpler sub-problems. Firstly, a binary classifier is run to identify customers with predicted CLTV greater than zero. Secondly, the customers predicted to shop again are split into five groups and independent regressors are trained for each group. 
% \todo[inline, color=green]{marc: would a diagram/illustration make sense here?}
% deep learning stuff
To the best of our knowledge, deep neural networks have not yet been successfully applied to the CLTV problem. However, there is work on the related churn problem (CLTV greater than zero) by~\citet{Wangperawonga} for telecommunications. The authors create a two-dimensional array for each customer where the columns are days of the year and the rows describe different kinds of communication (text, call etc.). They used this array to train deep convolutional neural networks. The model also used auto-encoders~\cite{vincent2008extracting} to learn low-dimensional representations of the customer arrays.

\subsection{Neural Embeddings}

% introduction to embeddings
Neural embeddings are a technique pioneered in Natural Language Processing (NLP) for learning distributed representations of words~\cite{Mikolov2013}. They provide an alternative to the one-of-$k$ (or one-hot) representation. Unlike the one-of-$k$ representation, which uses large sparse orthogonal vectors (which are equally similar), embeddings are compact, dense representations that encapsulate similarity. Embedded representations have been shown to give superior performance to sparse representations in several downstream tasks in language and graph analysis~\cite{Perozzi2014, Mikolov2013}. All embedding methods define a context that groups objects. Typically, the data is a sequence and the context is a fixed-length window that is passed along the sequence. The model learns that objects frequently occurring in the same context are similar and will have embeddings with high cosine similarity. 

% SGNS
One of the most popular embedding models is SkipGram with Negative Sampling (SGNS), which was developed by~\citet{Mikolov2013}. It has found a broad range of applications and we describe the most relevant to our work. \citet{Barkan2016} used SGNS for item-level embeddings in item-based collaborative filtering, which they called item2vec. As a context the authors use a basket of items that were purchased together \footnote{Items are synonymous with products, but the term `item' is used in the recommender systems literature.}. In~\cite{Grbovic2015}  SGNS was used to generate a set of product embeddings, which the authors called prod2vec, by mining receipts from email data. For each customer a sequence of products was built (with arbitrary ordering for those bought together) and then a context window was run over it. The goal was to predict products that are co-purchased by a single customer within a given number of purchases. The authors also proposed a hierarchical extension called bagged-prod2vec that groups products appearing together in a single email. \cite{Baeza-yates2015} used a variant of SGNS   to predict the next app a mobile phone user would open. The key idea is to consider sequences of app usage within mobile sessions. The data had associated time stamps, which the authors used to modify the original model. Instead of including every pairwise set of apps within the context, the selection probability was controlled by a Gaussian sampling scheme based on the inter-open duration.

\section{Customer Lifetime Value Model}

% introduction
The ASOS CLTV model uses a rich set of features to predict the net spend of customers over the next 12 months by training a random forest regressor on historic data. 
% problems with the target variable
One of the major challenges of predicting CLTV is the unusual distribution of the target variable. A large percentage of customers have a CLTV of zero. Of the customers with greater than zero CLTV, the values differ by several orders of magnitude. To manage this problem we explicitly model CLTV percentiles using a random forest regressor. Having predicted percentiles, the outputs are then mapped back to real value ranges for use in downstream tasks. 

% section overview
The remainder of this section describes the features used by the model, the architecture that allows the model to scale to over 10 million customers, and our training and evaluation methodology.

\subsection{Features}
\label{sec:model_features}

The model incorporates features from the full spectrum of customer information available at ASOS. There are four broad classes of data: (1) customer demographics, (2) purchase history (3) returns history (4) web and app session logs. By far the largest and richest of these classes are the session logs.

% features importance
We apply random forest feature importance \cite{hastie2001} to rank the 132 handcrafted features. Table~\ref{tab:featureimportanceclasses} shows the feature importance breakdown by the broad classes of data and Table~\ref{tab:featureimportance} shows the top features. As expected, the number of orders, the number of sessions in the last quarter and the nationality of the customer were very important features for CLTV prediction. However, we were surprised by the importance of the standard deviation of the order and session dates, particularly because the maximum spans of these variables were also features. We also did not expect the number of items purchased from the new collection to be one of the most relevant features. This is because newness is a major consideration for high value fashion customers.

\begin{table}[]
\centering
\caption{Feature importance by data class.}
\label{tab:featureimportanceclasses}
\begin{tabular}{r|c}
\textbf{Data class} & \textbf{Overall Importance}  \\
\hline
Customer demographics           & 0.078 \\
Purchases history          & 0.600 \\
Returns history           & 0.017 \\
Web/app session logs           & 0.345 \\
\end{tabular}
\end{table}

\begin{table}[]
\centering
\caption{Individual feature importance (top features).}
\label{tab:featureimportance}
\begin{tabular}{r|c}
\textbf{Feature Name} & \textbf{Importance}  \\
\hline
Number of orders               & 0.206 \\
Standard deviation of the order dates           & 0.115 \\
Number of session in the last quarter           & 0.114 \\
Country         & 0.064 \\
Number of items from new collection          & 0.055  \\
Number of items kept                & 0.049 \\
Net sales                   & 0.039 \\
Days between first and last session           & 0.039 \\
Number of sessions              & 0.035 \\
Customer tenure           & 0.033 \\
Total number of items ordered            & 0.025 \\
Days since last order    & 0.021 \\
Days since last session         & 0.019  \\
Standard deviation of the session dates          & 0.018 \\
Orders in last quarter        & 0.016 \\
Age                  & 0.014 \\
Average date of order        & 0.009 \\
Total ordered value                 & 0.008 \\
Number of products viewed                      & 0.007 \\
Days since first order in last year         & 0.006  \\
Average session date          & 0.006 \\
Number of sessions in previous quarter           & 0.005 \\
\end{tabular}
\end{table}

\subsection{Architecture}
\begin{figure}[tb]
\includegraphics[width=0.48\textwidth]{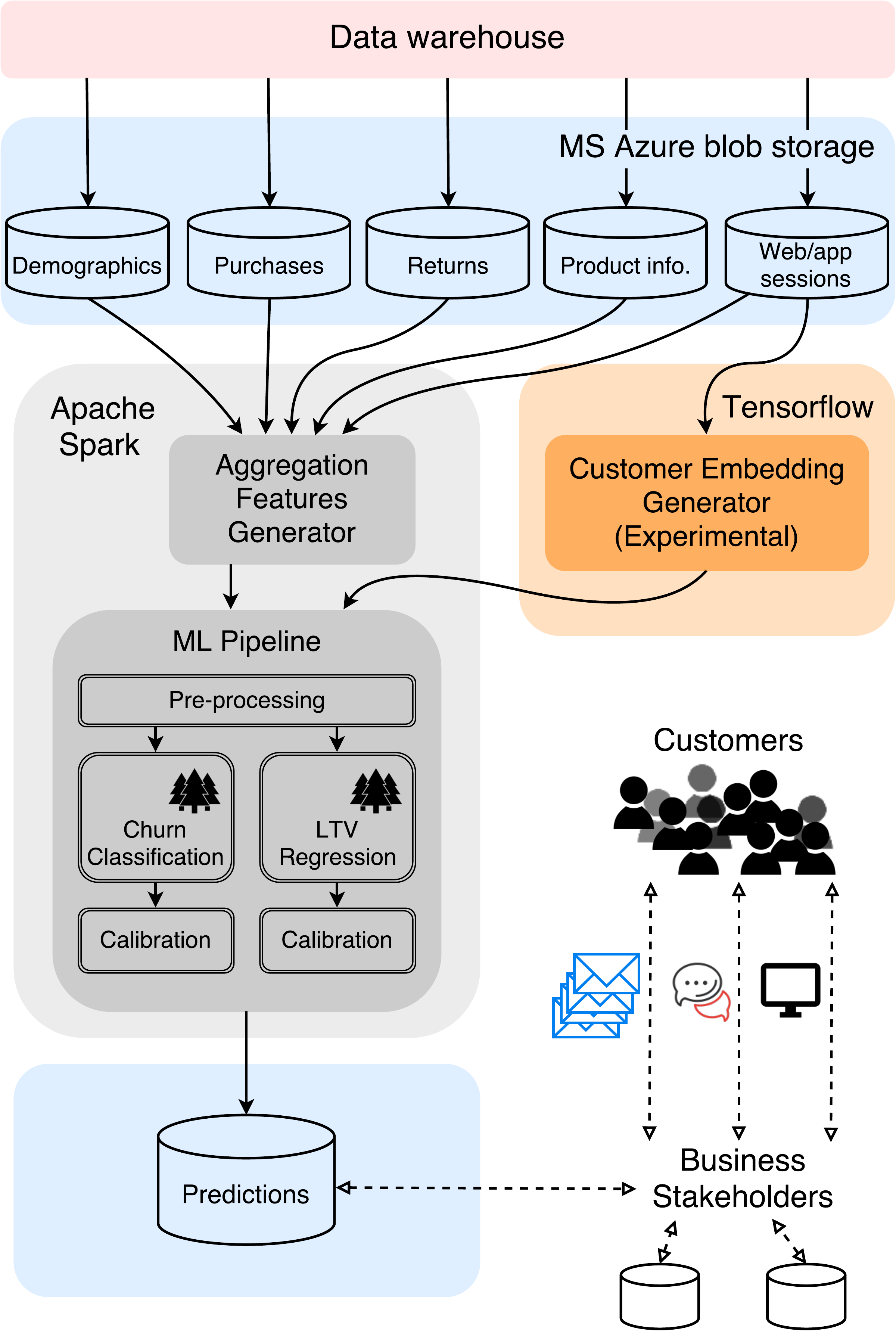}
\caption{High-level overview of the CLTV system. The solid arrows represent the flow of data, and the dashed arrows represents interaction between stakeholders and systems/data. Customer data is collected and pre-processed by our data warehouse and stored on Microsoft Azure blob storage. The processed data is used to generate handcrafted features in Spark clusters, with web/app sessions additionally used to produce experimental customer embeddings in Tensorflow. The handcrafted features and customer embeddings are then fed through the machine learning pipeline on Spark, which trains calibrated random forests for churn classification and CLTV regression. The resulting prediction are piped to operational systems.}
\label{fig:CLTVarchitecture}
\end{figure}

% High level overview of our CLTV system architecture/process
The high-level system architecture is shown in Figure~\ref{fig:CLTVarchitecture}. Raw customer data (see Section~\ref{sec:model_features}) is  pre-processed in our data warehouse and stored in Microsoft Azure blob storage\footnote{a Microsoft cloud solution that provides storage compatible with distributed processing using Apache Spark}. From blob storage, data flows through two work streams to generate customer features: A handcrafted feature generator on Apache Spark and an experimental customer embedding generator on GPU machines running Tensorflow\cite{Abadi2015}, which uses only the web/app sessions as input. The model is trained in two stages using an Apache Spark ML pipeline. The first stage pre-processes the features and trains random forests for churn classification and CLTV regression on percentiles. The second stage performs calibration and maps percentiles to real values. Finally the predictions are presented to a range of business systems that trigger personalised engagement strategies with ASOS customers.

\subsection{Training and Evaluation Process}
% live system
Our live system uses a calibrated random forest model with features from the past twelve months that is re-trained every day. We use twelve months because there are strong seasonality effects in retail and failing to use an exact number of years would cause fluctuations in features that are calculated by aggregating data over the training period. 

% training
To train the model we use historic net sales over the last year as a proxy for CLTV labels and learn the random forest parameters using features generated from a disjoint period prior to the label period. This is illustrated in Figure~\ref{fig:cltv_training}. Every day we generate:
\begin{enumerate}
   \item A set of aggregate features and product view-based embeddings for each customer  based on their demographics, purchases, returns and web/app sessions between two years and one year ago,
   \item Corresponding target labels, including the churn status and net one-year spend, for each customer based on data from the last year.
\end{enumerate}

The feature and label periods are disjoint to prevent information leakage.
% model evaluation
As the predictive accuracy of our live system could only be evaluated in a year's time, we establish our expectation on the performance of the model by forecasting for points in the past for which we already know the actual values as illustrated in Figure~\ref{fig:cltv_training}. We use the Area Under the receiver operating characteristic Curve (AUC) as a performance measure.
% We evaluate our trained random forest models as follows: We first select a subset of customers disjoint with the training set described above (as test set), and generate the set of features and target labels based on the same period as per training set. We then pass the features to the trained model, obtain the models' predicted labels based on the features, and compare with the actual labels to obtain the models' performance.

\subsection{Calibration}
% what is calibration
In this context, calibration refers to our efforts to ensure that the statistics of the model predictions are consistent with the statistics of the data. Model predictions are derived from RF leaf distributions and we perform calibration for both churn and CLTV prediction.

% calibration for churn
For customer churn prediction, choosing RF classifier parameters that maximise the AUC does not guarantee that the predictive probabilities will be consistent with the realised churn rate~\cite{zadrozny2001obtaining}. To generate consistent probabilities, we calibrate by learning a mapping between the estimates and the realised probabilities. This is done by training a one-dimensional logistic regression classifier to predict churn based only on the probabilities returned by the random forest. The logistic regression output is interpreted as a calibrated probability.
% calibration for CLTV
Similarly, to estimate CLTV we have no guarantees that the regression estimates achieved by minimizing the Root Mean Squared Error (RMSE) loss function will match the realised CLTV distribution. To address this problem, analogously to churn probability calibration, we first forecast the CLTV percentile and then map the predicted percentiles into monetary values. In this case, the mapping is learnt using a decision tree. We observe two advantages in performing calibration: (1) the model becomes more robust to the existence of outliers and (2) we obtain predictions, which when aggregated over a set of customers,  match the true values more accurately.

\subsection{Results}
\begin{figure}
\includegraphics[width=0.48\textwidth]{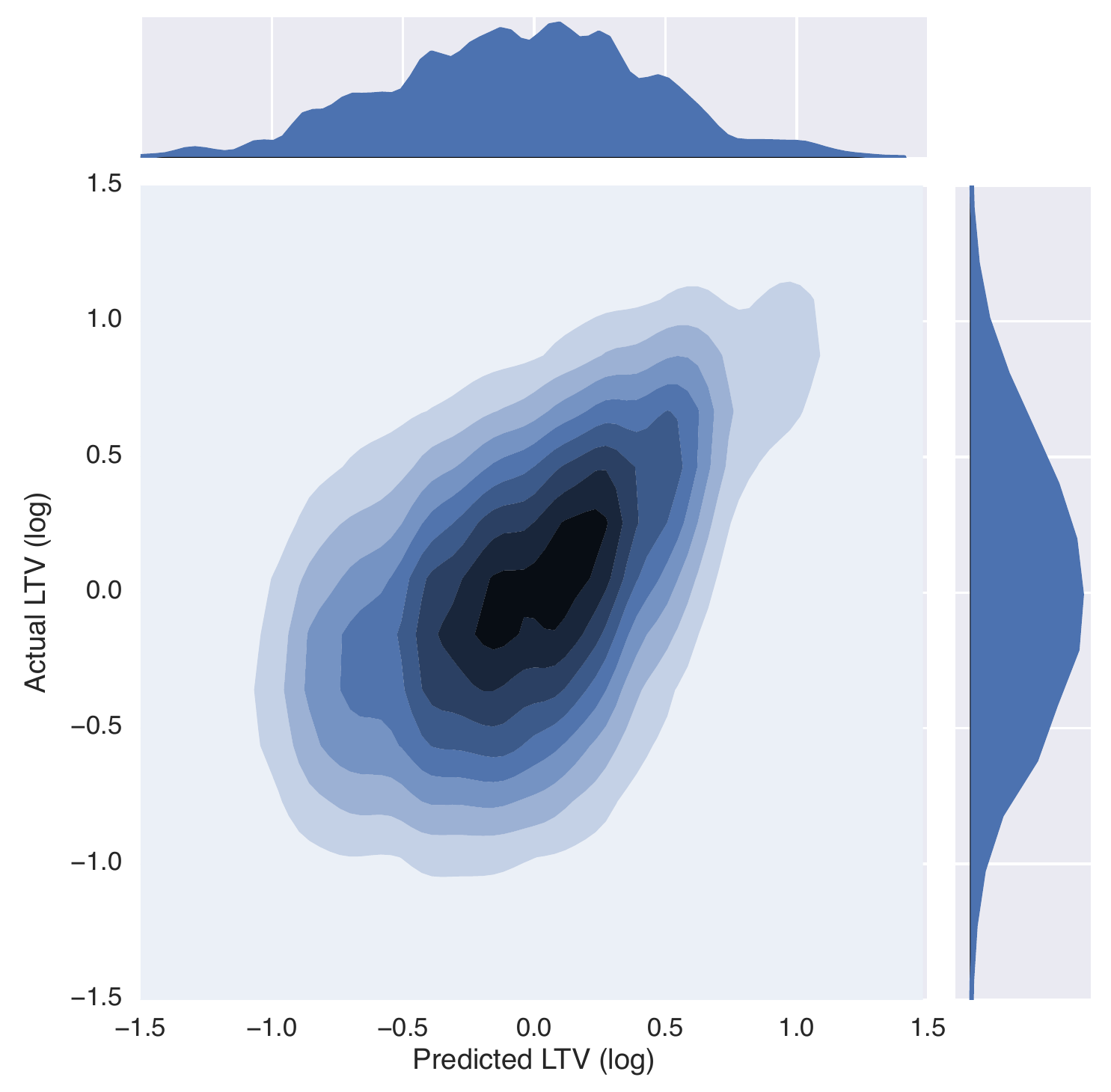}
\caption{Predicted CLTV against actual CLTV (excluding customers with an actual CLTV of 0). Units (horizontal and vertical axis) are the average CLTV value in GBP. The distribution of the prediction and the actual CLTV are similar in log scale (top and right density plots). The central plot shows the fit between the predictions and the actual values, which have a Spearman rank-order  correlation coefficient of 0.46. }
\label{fig:ltv_results}
\end{figure}

\begin{figure}[tb]
\includegraphics[width=0.48\textwidth]{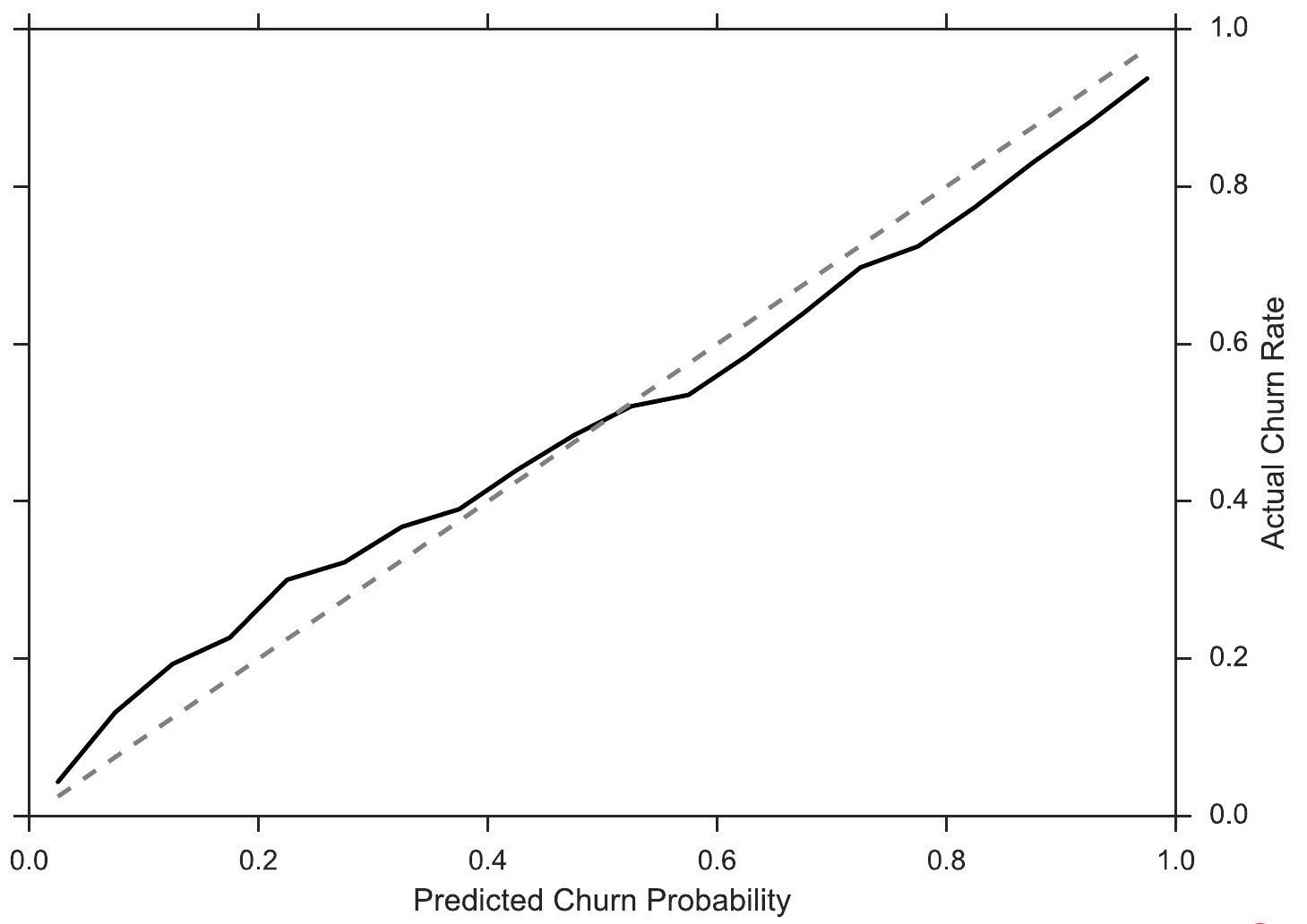}
\caption{Churn prediction density (horizontal axis) and match between predicted probabilities and actual probabilities (black line) versus the optimal calibration (dashed grey line). The predicted probabilities match closely with the actual probabilities}. 
\label{fig:churn_results}
\end{figure}

To find the optimal meta-parameters for the RFs we use 10-fold cross validation on a sample of the data.
For CLTV predictions (see Figure~\ref{fig:ltv_results}) 
we obtain Spearman rank-order correlation coefficients of 0.56 (for all customers) and 0.46 (excluding customers with a CLTV of 0). We can see in Figure~\ref{fig:ltv_results} that the range and density of the predicted CLTV matches the actual CLTV (excluding customers with an LTV of 0).
For the churn predictions (see Figure~\ref{fig:churn_results}) we  obtain an AUC of 0.798 and calibrated probabilities.

%\begin{figure}[tb]
%\includegraphics[width=0.48\textwidth]{ltvresiduals.pdf}
%\caption{LTV prediction residuals (prediction - actual). Units are relative to the actual average LTV value in monetary units. We can see that the model tends to slightly overestimate the LTV as the residuals are slightly shifted to the right.}
%\label{fig:ltv_residuals}
%\end{figure}

\section{Improving the CLTV model with feature learning}

In the remainder of the paper we describe our ongoing efforts to supplement the handcrafted features in our deployed system with automatic feature learning. Feature learning is the process of learning features directly from data, so as to maximise the objective function of the classification or regression task. This technique is frequently used within the realms of deep learning \citep{jia2014caffe} and dimensionality reduction \citep{roweis2000nonlinear} to overcome some of the limitations of engineered features. Despite being more difficult to interpret, learnt features avoid the resource intensive task of constructing features manually from raw data and have been shown to outperform the best handcrafted features in the domains of speech, vision and natural language. 
% In addition, embeddings project the data samples in a higher-dimensional space where they can be compared directly, allowing for a number of other applications, including clustering and user/customer segmentation. 

We experiment with two distinct approaches. Firstly, we learn unsupervised neural embeddings using customer product views. Once learnt the embeddings are added to the feature set of the RF model. Secondly, we train a hybrid model that combines logistic regression with a Deep Neural Network (DNN). The DNN uses the handcrafted features to learn higher order feature representations.

\subsection{Embedding Customers using Browsing Sessions}
\label{sec:embcustsessions}
\begin{figure}[tb]
\includegraphics[width=0.48\textwidth]{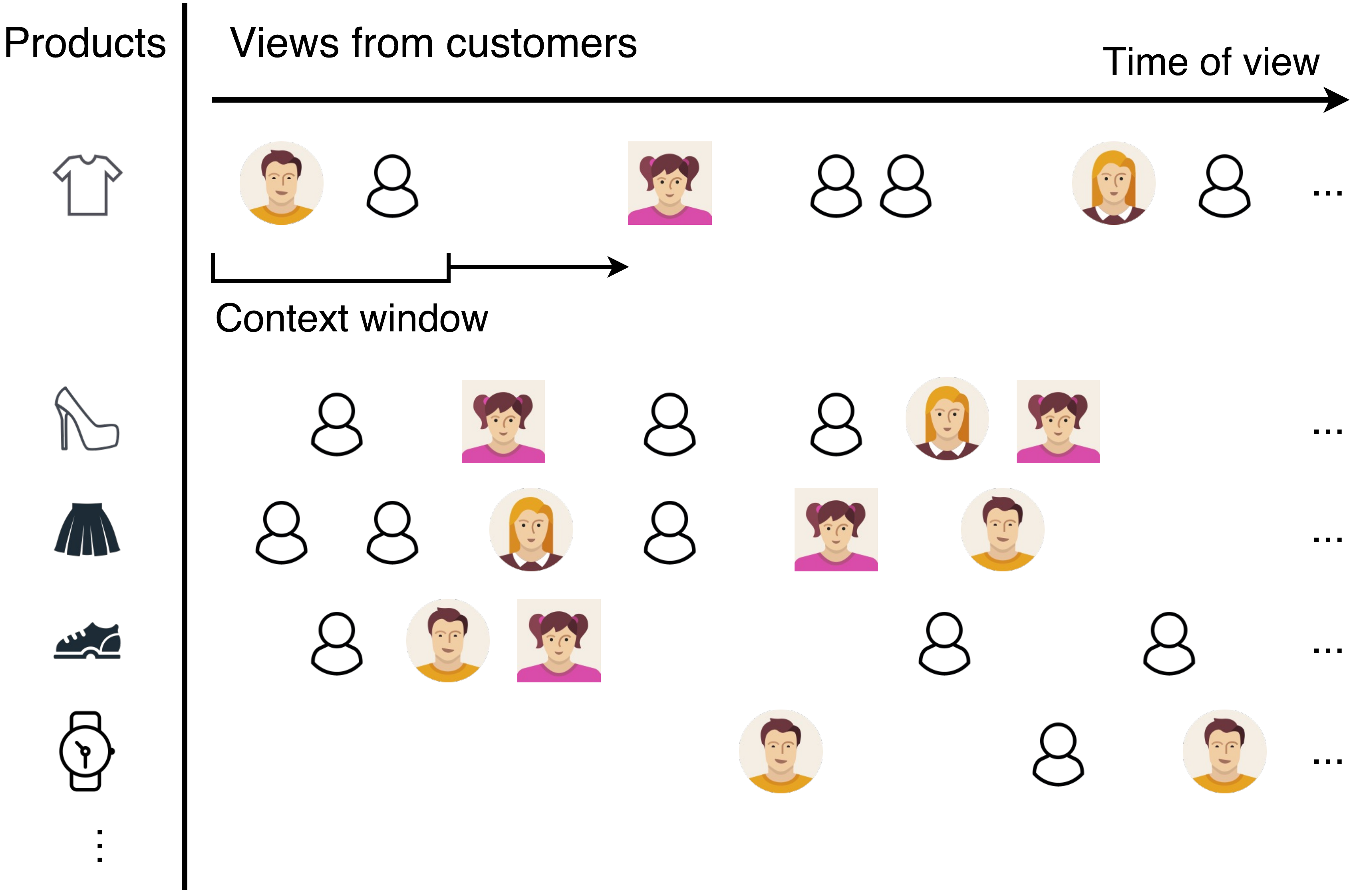}
\caption{Customer pair generation in the skip-gram model based on product views. Each row represents a product sold on ASOS and the sequence of customer views of that product. In this example the context window is of length two and considers only adjacent view events (each pair in this example) of the same product. Hence, the exact time a product is viewed is ignored. Customers who often appear  in the same context window will be close to each other in the embedding representation.}
\label{fig:customer_embedding_context_window}
\end{figure}

% why we need embeddings
We learn embeddings of ASOS customers using neural embedding models borrowed from NLP. To re-purpose them  we replace sequences of words with sequences of customers viewing a product (see Figure~\ref{fig:customer_embedding_context_window}). 
% justification for this approach
Previous work has looked at embedding products based on sequences of customer interactions~\cite{Grbovic2015, Vasile2016a, Barkan2016}. It is possible to aggregate product embeddings to produce a customer embedding. However, this approach fails at the task of producing long-term forecasts when products are relatively short live (as is the case in the fashion industry). For this reason we learn embeddings of customers directly.
% intuitive justification
Intuitively, high-value customers tend to browse products of higher value, less popular products and products that may not be at the lowest price on the market. By contrast, lower-value customers will tend to appear together in product sequences during sales periods or for products that are priced below the market. This information is difficult to incorporate into the model using hand-crafted features as the number of sequences of product views grows combinatorially.

\begin{figure}[tb]
\includegraphics[width=0.48\textwidth]{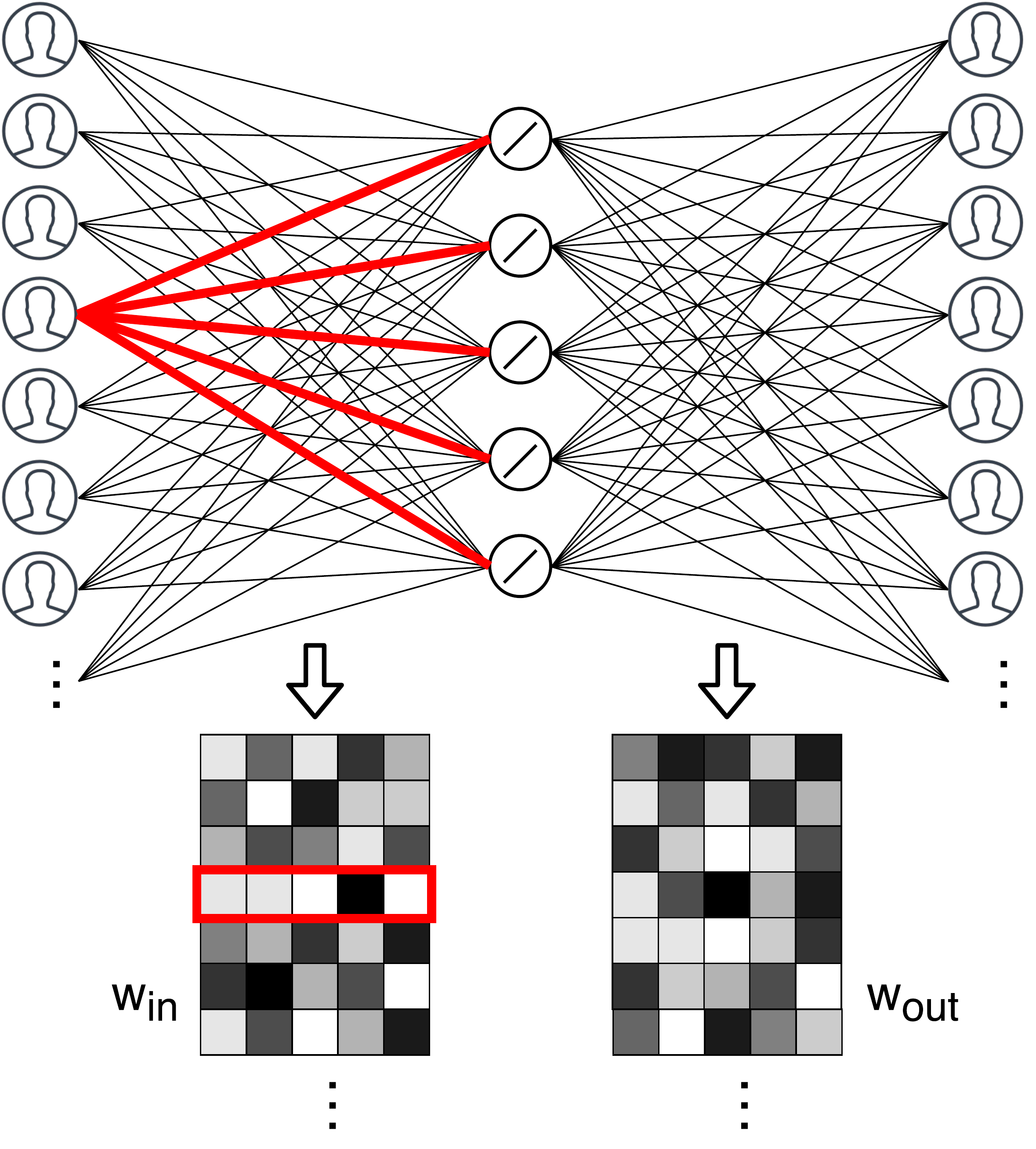}
\caption{Neural network architecture and matrix representation of the network's input/output weights (embedding representation) in the Skipgram model on customer embeddings. The Skipgram model uses a neural network with one hidden layer.
Customers are represented by one-hot vectors in the input and output layers, and the weights between the input (output) and hidden layers are represented by a randomly initialised customer input (output) embedding matrix $\mathbf{W}_{\textrm{in}}$ ($\mathbf{W}_{\textrm{out}}$). Each row of the matrix represents a customer embedding in $\mathbb{R}^n$, with lighter cells representing a more negative value and darker cells representing a more positive value.}
%\todo[inline]{Include the following? If not this todo can be safely deleted, "For each training example applied sequentially, the dot product between the input customer and all (or some in the case of Skipgram with negative sampling (SGNS)) output customers are computed, with result passed though a softmax function and compared with the one-hot vector representing the output customer. The error is then back-propagated to all (or some in SGNS) connections in the network."}
%\todo[inline, color=green]{marc: Explain how the dot-product is related to angles between vectors, which then allows us to say something about similarity between the vectors. Give the geometric motivation.}
\label{fig:customer_skipgram_model}
\end{figure}

% describe the basic embedding model
Figure~\ref{fig:customer_skipgram_model} shows the neural architecture of our customer embedding model. 
The model has two large weight matrices, $\mathbf{W}_{\rm in}$ and $\mathbf{W}_{\rm out}$ that learn distributed representations of customers. The output of the model is $\mathbf{W}_{\rm in}$ and after training, each row of $\mathbf{W}_{\rm in}$ is the vector representation of a customer in the embedding space. The inputs to the model are pairs of customers $(C_{\rm in}, C_{\rm out})$ and the loss function is the probability of observing the output customer $C_{\rm out}$ given $C_{\rm in}$:
\begin{equation}
E = -\log P(C_{\rm out} | C_{\rm in}) = \frac{{\rm exp}\left( {\mathbf{v}^{\prime}_{\rm out}}^{T} \mathbf{v}_{\rm in} \right)}{\sum_{j=1}^{|C|} {\rm exp}\left( {\mathbf{v}^{\prime}_{j}}^{T} \mathbf{v}_{\rm in}  \right)} ,
\label{eq:loss_function}
\end{equation}
where ${\mathbf{v}^{\prime}_j}$ represents the $j^{\rm th}$ row of $\mathbf{W}_{\rm out}$, $v^{\prime}_{\rm out}$ represents the row of $W_{\rm out}$ that corresponds to the customer $C_{\rm out}$ and $v_{\rm in}$ represents the row of $W_{\rm in}$ that corresponds to the customer $C_{\rm in}$ and $|C|$ is the total number of customers.

% negative sampling
The output softmax layer in Figure~\ref{fig:customer_skipgram_model} has a unit for every customer, which must be evaluated for each training pair. This would be prohibitively expensive to do for approximately ten million customers. However, it can be approximated using SkipGram with Negative Sampling (SGNS) by only evaluating a small random selection of the total customers at each training step~\cite{Mikolov2013}. 

Applying SGNS requires three key design decisions: 
\begin{enumerate}
\item How to define a context.
\item How to generate pairs of customers $(C_{\rm in}, C_{\rm out})$ from within the context.
\item How to generate negative samples.
\end{enumerate}

%\todo[inline, color=green]{fix next sentence}
In NLP, the context is usually a sliding window over word sequences within a document. The word at the centre of the window is the input word and $(\text{word}_{\rm in}, \text{word}_{\rm out})$ pairs are formed with every other word in the context. The negative samples are drawn from a modified unigram distribution of the training corpus\footnote{`Modified' means that word frequencies are raised to a power before normalisation (usually 0.75).}. We adopt this approach here. 

Figure~\ref{fig:customer_embedding_context_window} shows how we define a context and generate customer pairs. Each product in the catalogue is associated with a sequence of customer views. A sliding context window is then passed over the sequence of customers. For every position of the context window, the central customer is used as $C_{\rm in}$ and all other customers in the window are used to form $(C_{\rm in}, C_{\rm out})$ pairs. In this way, a window of length three containing $(C_1, C_2, C_3)$ would generate customer pairs $(C_2, C_1)$ and $(C_2, C_3)$. We empirically found that a window of length 11 worked well. 

%We also experimented with embedding products and aggregating the vectors of products (sum or mean) viewed by a customer to produce customer embeddings. We found that directly learning customer embeddings gave improved CLTV predictive performance. For the negative sampling distribution we used the total number of product views by each customer in the training set raised to 0.75.

% when applied to words are IID word pairs. The model employs two separate vector representations, one for the first (input) word and another for the second (context) word. \todo[inline]{link to figure} The word pairs are generated by taking a sequence of words and running a sliding window (the context) over them. As an example the word sequence "chance favours the prepared mind" with a window of size 3 would generate the following training data \{(chance, favours), (chance, the), (favours, chance), (favours, the), (favours, prepared), (the, chance), (the, favours), (the, prepared), (the, mind), (prepared, favours), (prepared, the), (prepared, mind), (mind, the), (mind, prepared)\}.

The embedding algorithm begins by randomly initialising $\mathbf{W}_{\rm in}$ and $\mathbf{W}_{\rm out}$. Then, for each customer pair $(C_{\rm in}, C_{\rm out})$ with embedded representations $(\mathbf{v}_{\rm in}, \mathbf{v}^{\prime}_{\rm out})$, $k$ negative customer samples $C_{\rm neg}$ are drawn. After the forward pass, $k+1$ rows of $\mathbf{W}_{\rm out}$ are updated via gradient descent, using backpropagation:

\begin{align}
{\mathbf{v}^{\prime}_j}^{\rm new} &= 
\begin{cases} 
{\mathbf{v}^{\prime}_j}^{\rm old} - \eta\,(\sigma({\mathbf{v}^{\prime}_j}^{T}\mathbf{v}_{\rm in}) - t_j)\,\mathbf{v_{\rm in}} & \forall j: \ C_j \in C_{\rm out} \cup C_{\rm neg}\\ 
{\mathbf{v}^{\prime}_j}^{\rm old} & \text{otherwise} \quad ,
\end{cases}
\label{eq:wout update}
\end{align}
where $\eta$ is the update rate, $\sigma$ is the logistic sigmoid and $t_j = 1$ if $C_{\rm in} = C_{\rm out}$ and $0$ otherwise.

Finally, only one row of $\mathbf{W}_{\rm in}$, corresponding to $\mathbf{v}_{\rm in}$ is updated according to:
\begin{align}
\mathbf{v}_{\rm in}^{\rm new} &= 
\mathbf{v}_{\rm in}^{\rm old} - \eta \sum_{j :\, {C}_j \in C_{\rm out} \cup C_{\rm neg}} (\sigma({\mathbf{v}^{\prime}_j}^{T} \mathbf{v}_{\rm in}) - t_j)\,\mathbf{\mathbf{v}^{\prime}_j} \quad .
\label{eq:win update}
\end{align}

\begin{figure}
\includegraphics[width=0.48\textwidth]{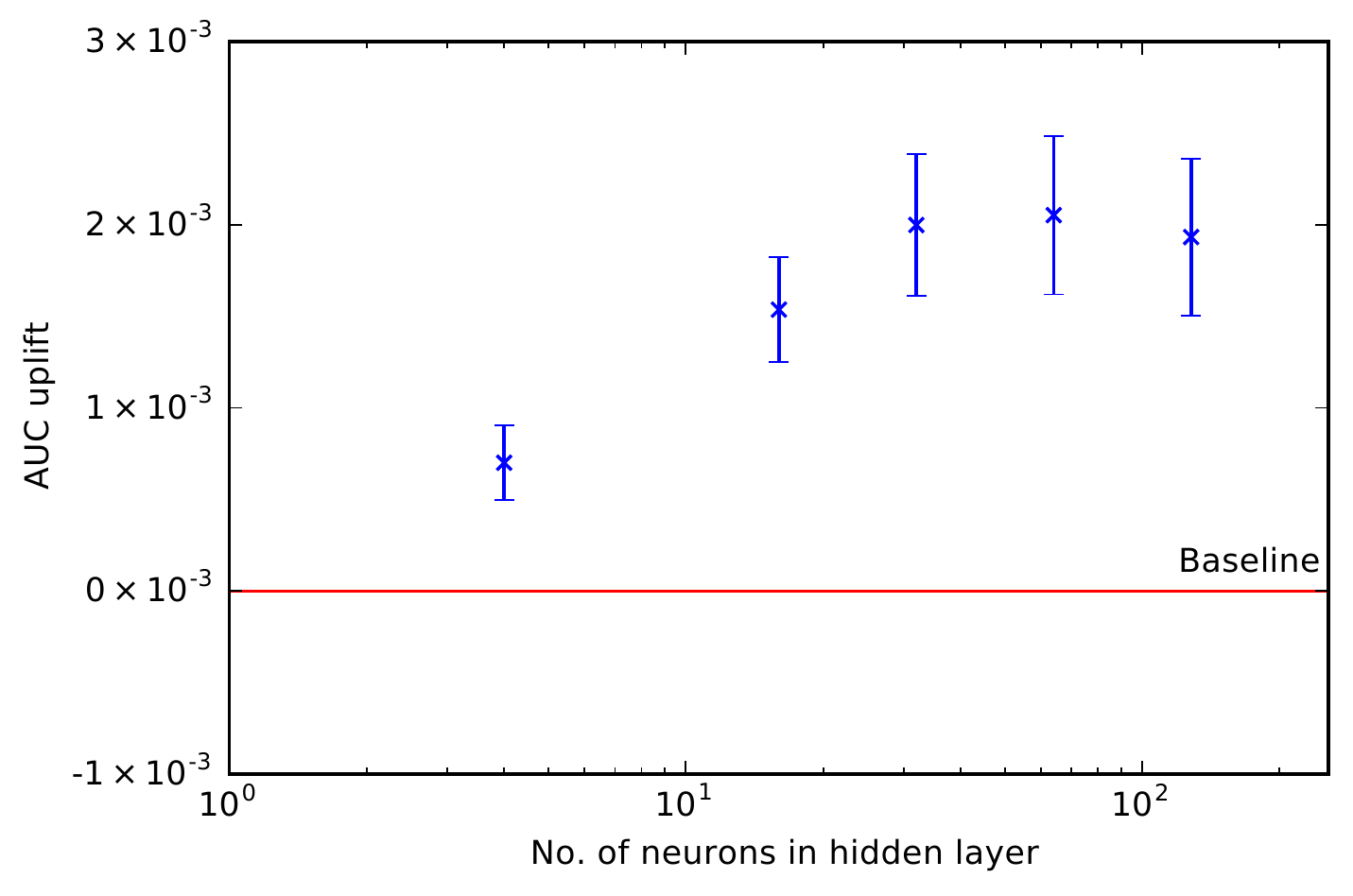}
\caption{Uplift in the area under the receiver operating characteristics curve achieved on random test sets of 20,000 customers with product view-based embeddings against number of neurons in the hidden layer of the Skipgram model. The error bars represent the 95\% confidence interval of the sample mean.}
\label{fig:embedding_auc_uplift}
\end{figure}

Figure~\ref{fig:embedding_auc_uplift} shows that we obtained a significant uplift in AUC using embeddings of customers. We experimented with a range of embedding dimensions and found the best performance to be in the range 32-128 dimensions. This result shows that this approach is highly relevant and we are working to incorporate the technique into our live system at the point of writing this paper.

% \begin{align}
% v_j^{'new} &= 
% \begin{cases} 
% (v_j^{'old} - \eta(\sigma(v_j^{'T}v_I) - t_j)\mathbf{v_I}, & \forall j: \ C_j \in C_O \cup C_{neg}\\ 
% v_j^{'old}, & \text{otherwise} 
% \end{cases}
% \end{align}
% where $\eta$ is the update rate, $\sigma$ is the logistic sigmoid and $t_j = 1$ if $v_j = v_O$ and 0 otherwise. Finally, only the one row of $W_{in}$ corresponding to $v_I$ is updated according to
% \begin{align}
% v_I^{new} &= 
% v_I^{old} - \eta \sum_{j : w_j \in w_O \cup W_{neg}} (\sigma(v^{'T}_jv_I) - t_j)\mathbf{v'_j} 
% \end{align}

\subsubsection{Embeddings for the Live System}
% problem: embeddings dimensions switch
To use embeddings in the deployed system it is necessary to make a correspondence between the embedding dimensions in the training period and the live system's feature generation period. Figure~\ref{fig:cltv_training} shows that the features for training the CLTV model and features used for the live system come from disjoint time periods. As the embedding dimensions are unlabelled, randomly initialised and exchangeable in the SGNS loss function (see Equation~\eqref{eq:loss_function}), the parameters learned in the training period can not be assumed to match the embeddings used in the live system. 
% solution: initialisation
To solve this problem, instead of randomly initialising $W_{in}$ and $W_{out}$ we perform the following initialisation:
\begin{itemize}
\item Customers that were present in the training period: initialise with training embeddings.
\item New customers: initialise to uniform random values with absolute values that are small compared to the training embeddings.
\end{itemize}
% sorting pairs
In the live system there are four types of $(C_{\rm in}, C_{\rm out})$ customer pairs: $(C_{\rm new}, C_{\rm new}), (C_{\rm new}, C_{\rm old}), (C_{\rm old}, C_{\rm new})$ and $(C_{\rm old}, C_{\rm old})$. Equation~\eqref{eq:win update} shows that the update to $v_{\rm in}$ is a linear combination of $v_{\rm out}$ and the negative vectors. Therefore a single update of a $(C_{\rm new}, C_{\rm old})$ pair is guaranteed to be a linear combination of embedding vectors from the training period. To generate embeddings that are consistent across the two time periods we order the data in each training epoch as  (1) $(C_{\rm old}, C_{\rm old})$ to update the  representation of the old customers, (2) $(C_{\rm new}, C_{\rm old})$ to initialise the new customers with linear combinations of embeddings from old customers, (3) $(C_{\rm old}, C_{\rm new})$, (4) $(C_{\rm new}, C_{\rm new})$. 
% embedding customers rather than products
For this scheme to work there must be a large proportion of customers that are present in both the training and test periods. This is true for customer embeddings \textbf{but it is not true for product embeddings}. This requirement explains why we chose to learn customer embeddings directly instead of using aggregations of product embeddings.

% \todo[inline]{angelo: moved this from discussion}
% There are some key difference between words in a corpus and products in catalog that motivate a different approach to the embedding problem than the vanilla SGNS model:
% \begin{enumerate}
% \item products come with a time stamp
% \item products are ephemeral - they sell out and new products appear
% \item products have both substitutes and complements, which are often viewed together
% \item products have hierarchical groupings where the products that a customer views can be grouped into web sessions
% \end{enumerate}

\subsection{Embeddings of Handcrafted Features}
\label{sec:embhandcraftfeat}
% Introduction: if deep network will improve performance

We also investigate to what extent our deployed system could be improved by replacing the RF with a Deep Neural Network (DNN). This is motivated by the recent successes of DNNs in vision, speech recognition, and recommendation systems~\cite{lecun2015deep, Covington2016}. Our results indicate that while incorporating DNNs in our system may improve the performance, the monetary cost of training the model outweighs the performance benefits.

% Begin (explore feasibility) by looking at churn first
We limit our experiments with DNNs to the churn problem (predicting a CLTV of zero for the next 12 months). This reduces the CLTV regression problem to a binary classification problem  with more interpretable predictions and metrics on performance.

% Describe details of network architecture/training being used in experiments
We experiment with (1) deep feed-forward neural networks and (2) hybrid models combining logistic regression and a deep feed-forward neural network similar to that used in~\cite{Cheng2016a}. The deep feed-forward neural networks accept all continuous-valued features and dense embeddings of categorical features, which are described in Section~\ref{sec:model_features}, as inputs. We use Rectified Linear Units (ReLU) activations in the hidden units and sigmoid activation in the output unit. The hybrid models are logistic regression models incorporating a deep feed-forward neural network. The output of the neural network's final hidden layer is used alongside all continuous-valued features and sparse categorical features, as input. This is akin to skip connections in neural networks described in \cite{raiko2012}, with the inputs connected directly to the output instead of the next hidden layer. Training on the neural network part of the models is done via mini-batch stochastic gradient descent with Adagrad optimiser, with change of weights backpropagated to all layers of the network. Regularisation is applied on the logistic regression part of the hybrid models via the use of FTRL-Proximal algorithm as described by \citet{McMahan2013}.

% Describe our experiment and what metrics we are collecting
We evaluate the performance and scalability of the two models with different architectures, and compare them with other machine learning techniques. We experimented with neural networks with two, three and four hidden layers, each with different combinations of number of neurons. For each neuron architecture, we train the models using a subset of customer data (the training set), and record (1) the maximum AUC achieved when evaluating on a separate subset of customer data (the test set), (2) the (wall clock) time taken to complete a pre-specified number of training steps. We repeat the training/testing multiple times for each architecture to obtain an estimate on the maximum AUC achieved and the training time. All training/testing is implemented using the TensorFlow library~\cite{Abadi2015} on a Tesla K80 GPU machine.

% Hybrid (wide & deep) model outperforms deep only model
\begin{figure}[tb]
\includegraphics[width=0.48\textwidth]{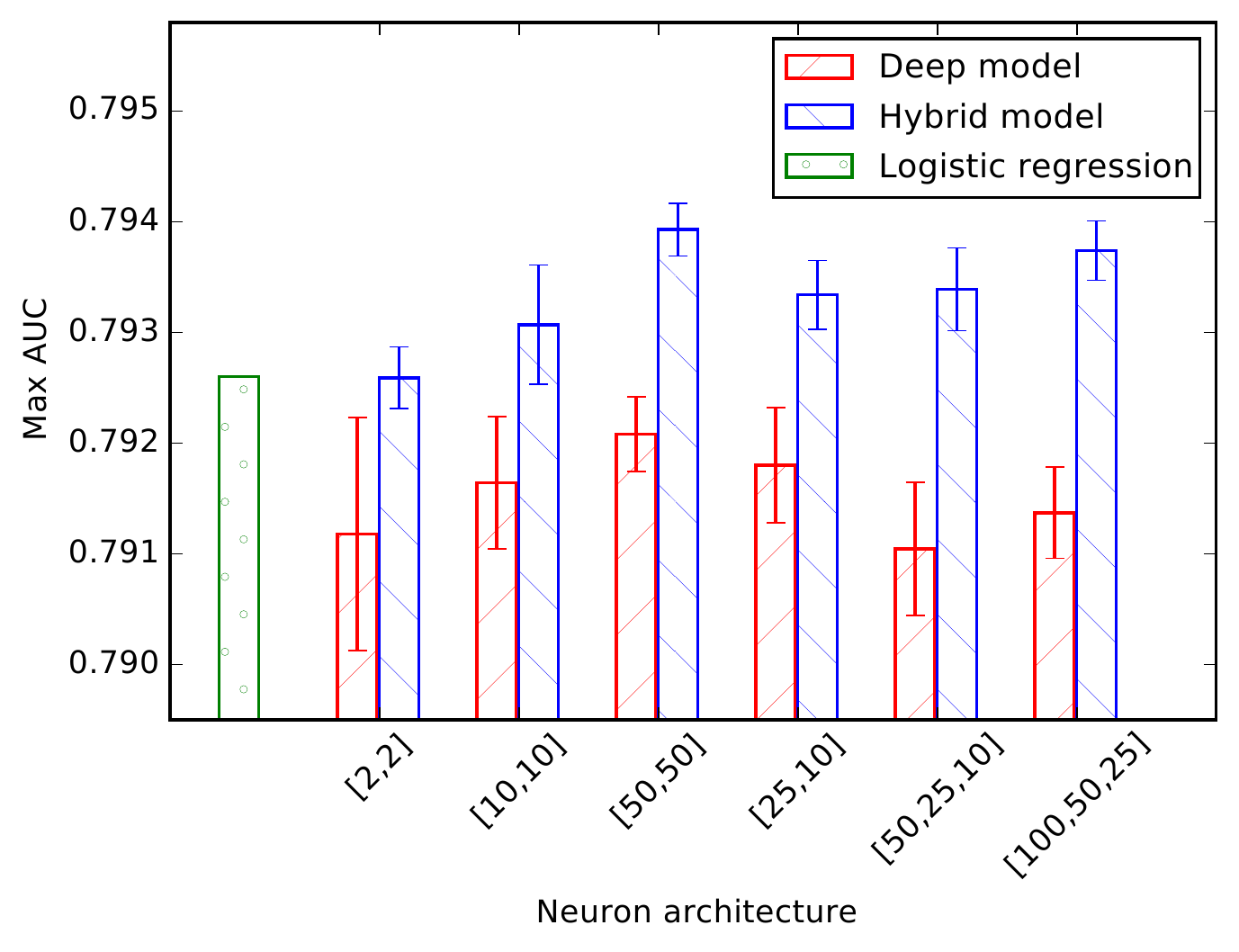}
\caption{Maximum area under the receiver operating characteristics curve achieved on a test set of 50,000 customers in deep feed-forward neural networks and hybrid models with different numbers of hidden layer neurons. The error bars represent the 95\% confidence interval of the sample mean. The number of hidden-layer neurons are recorded in the following format: $[x, y]$ denotes a neural network with $x$ and $y$ neurons in the first and second hidden layer respectively, $[x, y, z]$ denotes a neural network with $x$, $y$, $z$ neurons in the first, second, and third hidden layers.}
\label{fig:deepvshybridAUC}
\end{figure}

Introducing bypass connections in the hybrid models improves the predictive performance compared to a deep feed-forward neural network with the same architecture. Figure~\ref{fig:deepvshybridAUC} shows a comparison of the maximum AUC achieved by DNNs to the hybrid logistic and DNN models on a test set of 50,000 customers. Our experiments show a statistically significant uplift at least $1.4 \times 10^{-3}$ in every configuration of neurons that we experimented with. We believe the uplift is due to the hybrid models' ability to memorize the relationship between a set of customer attributes and their churn status in the logistic regression part. This is complementary to a deep neural network's ability to generalise customers based on other customers who have similar aggregation features, as described by \citet{Cheng2016a}.

% Comparison between hybrid, logistic regression and random forest
\begin{figure}[tb]
\centering
\includegraphics[width=0.48\textwidth]{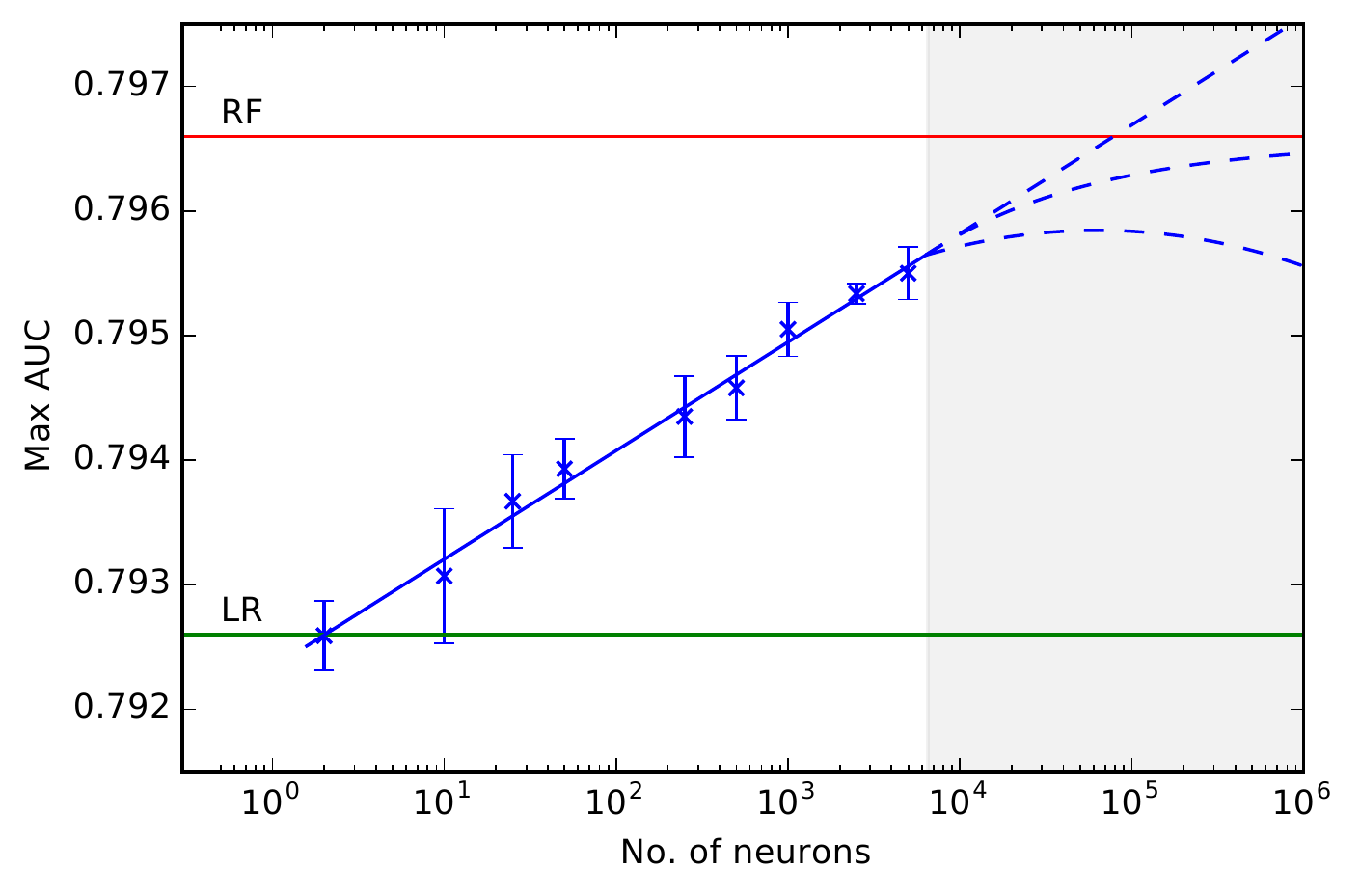}
\caption{Maximum Area Under the receiver operating characterics Curve (AUC) achieved on a test set of 50,000 customers in hybrid models against number of neurons in the hidden layers (in log scale). We only consider hybrid models with two hidden layers, each having the same number of neurons. The error bars represent the 95\% confidence interval of the sample mean. The bottom (green) and top (red) horizontal line represent the maximum AUC achieved by a vanilla logistic regression model (LR) and our random forest model (RF) on the same set of customers. The dashed lines in the shaded region represent different forecast scenarios for larger architectures.}
\label{fig:hybridAUCvsnoneurons}
\end{figure}

We tried to estimate the size (in number of neurons) of hybrid model required to outperform the AUC of the RF model. Figure~\ref{fig:hybridAUCvsnoneurons} shows that there is a linear relationship between the maximum AUC achieved on the same test set of customers and the number of neurons in each hidden layer in logarithmic scale. We notice a hybrid model with a small number of neurons in each hidden layer already gives statistically significant improvement in maximum AUC achieved compared with a vanilla logistic regression~\footnote{A vanilla LR is essentially a hybrid model minus the deep network part, with the same input and optimisation wherever relevant.} (LR), but within the range of our experiments we could not exceed the performance of the RF model. The shaded regions and dashed lines in Figure~\ref{fig:hybridAUCvsnoneurons} provide three estimates of the number of neurons that would be required to outperform the RF model.

% Hybrid might outperform random forest, but training cost not justified
\begin{figure}[tb]
\includegraphics[width=0.48\textwidth]{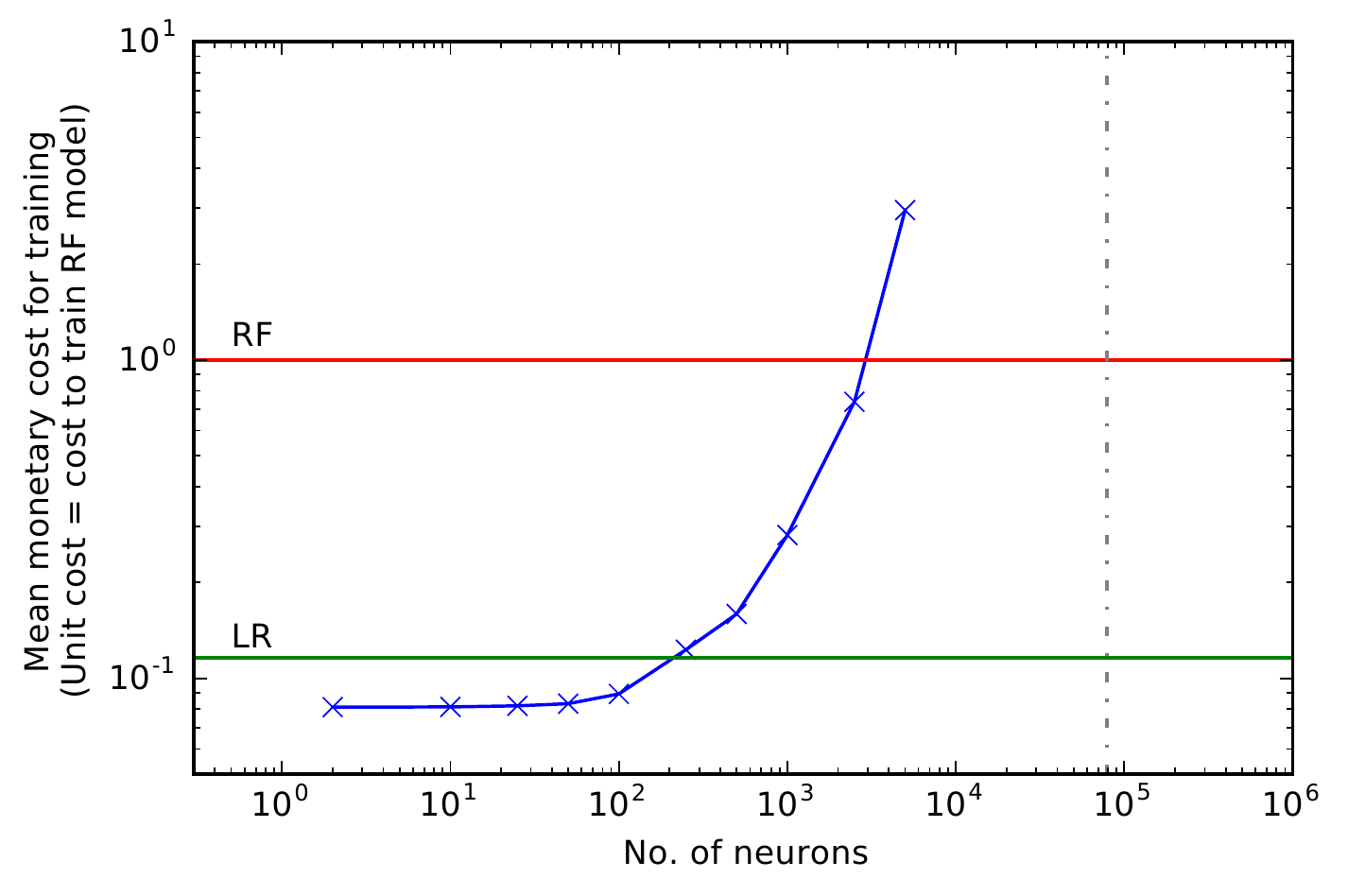}
\caption{Mean monetary cost to train hybrid models on a training set of 100,000 customers against the number of neurons in the hidden layers (both in log scale). The training cost shown is relative to the cost of training our random forest (RF) model. Here we only consider hybrid models with two hidden layers, each having the same number of neurons. The bottom (green) and top (red) horizontal line represents the mean cost to train a vanilla logistic regression model (LR) and our RF model on the same set of customers. The cost shown is based on the time required to train each of the models, and the cost of using the computational resources - Spark clusters to train RF models, and GPU VMs to train LR and hybrid models in Microsoft Azure. The vertical dash-dotted (grey) line represents the estimated number of neurons in each layer required for a two-hidden layer hybrid model to out-perform our random forest model.}
\label{fig:hybridRuntimeCost} 
\end{figure}
While the experiments suggest it is possible for our hybrid model, which incorporates a deep neural network, to outperform the calibrated RF model in churn classification, we believe the monetary cost required to perform such training outweighs the benefit of gain in performance. 
Figure~\ref{fig:hybridRuntimeCost} shows the relationship between monetary cost to train the hybrid models and the number of neurons in each hidden layer. The cost of training a vanilla LR model and our calibrated RF model are indicated by horizontal lines. The cost rises exponentially with increasing number of neurons, indicating that a hybrid model that outperformed our calibrated RF model is not be practical on cost grounds.

% extension to CLTV
We believe the case is similar for CLTV prediction, in which a hybrid model on handcrafted features (whether in numerical values or categorical feature embeddings) can achieve better performance than our current deployed random forest model, though with a much higher cost that is not commercially viable. This is supported in principle by our preliminary experiments, which measures the root mean squared error (RMSE) between the hybrid models' predicted percentile and actual percentile of each customer's spend. We observe increasing the number of neurons in hybrid models decreases the RMSE, but are unable to train hybrid models with tens of thousands of neurons due to prohibitive runtimes.

\section{Discussion and Conclusions}
% summary of the first half
We have described the CLTV system deployed at ASOS and the main issues we faced while building it. In the first half of the paper we describe our baseline architecture, which achieves state of the art performance for this problem and discuss an important issue that is often overlooked in the literature: model calibration.
% summary of second half
Given the recent success of representation learning across a wide range of domains, in the second half of the paper we focus on our ongoing efforts to further improve the model by learning two additional types of representations: Training feedforward neural network on the handcrafted features in a supervised setting (Section \ref{sec:embhandcraftfeat});  by learning an  embedding of customers  using session data in an unsupervised setting to augment our set of RF features (Section \ref{sec:embcustsessions}).
% embeddings worked
We showed that learning an embedding of a rich source of data (products viewed by a customer) in an unsupervised setting can improve the performance over using only handcrafted features and plan to incorporate this embedding in the live system as well as apply this same approach to other types of events (e.g. products bought by a customer).
% future work
The main alternative approach to the two described ways of learning representations would be to use a deep network to learn end-to-end from raw data sources as opposed to using handcrafted features as inputs. We are starting to explore this approach, which while extremely challenging, we believe might also provide large improvement versus the state of the art.

\section*{Acknowledgements}
This work was partly funded by an Industrial Fellowship from the Royal Commission for the Exhibition of 1851. The authors thank Jedidiah Francis for useful discussions and the anonymous reviewers for providing many improvements to the original manuscript.

% \appendix
% Appendix A
% \section{Headings in Appendices}

\bibliographystyle{ACM-Reference-Format}
\balance
%\bibliography{Mendeley_ASOS,otherrefs} 
\bibliography{otherrefs}

%%% -*-BibTeX-*-
%%% Do NOT edit. File created by BibTeX with style
%%% ACM-Reference-Format-Journals [18-Jan-2012].

\begin{thebibliography}{00}

%%% ====================================================================
%%% NOTE TO THE USER: you can override these defaults by providing
%%% customized versions of any of these macros before the \bibliography
%%% command.  Each of them MUST provide its own final punctuation,
%%% except for \shownote{}, \showDOI{}, and \showURL{}.  The latter two
%%% do not use final punctuation, in order to avoid confusing it with
%%% the Web address.
%%%
%%% To suppress output of a particular field, define its macro to expand
%%% to an empty string, or better, \unskip, like this:
%%%
%%% \newcommand{\showDOI}[1]{\unskip}   % LaTeX syntax
%%%
%%% \def \showDOI #1{\unskip}           % plain TeX syntax
%%%
%%% ====================================================================

\ifx \showCODEN    \undefined \def \showCODEN     #1{\unskip}     \fi
\ifx \showDOI      \undefined \def \showDOI       #1{{\tt DOI:}\penalty0{#1}\ }
  \fi
\ifx \showISBNx    \undefined \def \showISBNx     #1{\unskip}     \fi
\ifx \showISBNxiii \undefined \def \showISBNxiii  #1{\unskip}     \fi
\ifx \showISSN     \undefined \def \showISSN      #1{\unskip}     \fi
\ifx \showLCCN     \undefined \def \showLCCN      #1{\unskip}     \fi
\ifx \shownote     \undefined \def \shownote      #1{#1}          \fi
\ifx \showarticletitle \undefined \def \showarticletitle #1{#1}   \fi
\ifx \showURL      \undefined \def \showURL       #1{#1}          \fi
% The following commands are used for tagged output and should be
% invisible to TeX
\providecommand\bibfield[2]{#2}
\providecommand\bibinfo[2]{#2}
\providecommand\natexlab[1]{#1}

\bibitem[\protect\citeauthoryear{Abadi, Agarwal, Barham, Brevdo, Chen, Citro,
  Corrado, Davis, Dean, Devin, Ghemawat, Goodfellow, Harp, Irving, Isard, Jia,
  Kaiser, Kudlur, Levenberg, Man, Monga, Moore, Murray, Shlens, Steiner,
  Sutskever, Tucker, Vanhoucke, Vasudevan, Vinyals, Warden, Wicke, Yu, and
  Zheng}{Abadi et~al\mbox{.}}{2015}]%
        {Abadi2015}
\bibfield{author}{\bibinfo{person}{Martin Abadi}, \bibinfo{person}{Ashish
  Agarwal}, \bibinfo{person}{Paul Barham}, \bibinfo{person}{Eugene Brevdo},
  \bibinfo{person}{Zhifeng Chen}, \bibinfo{person}{Craig Citro},
  \bibinfo{person}{Greg Corrado}, \bibinfo{person}{Andy Davis},
  \bibinfo{person}{Jeffrey Dean}, \bibinfo{person}{Matthieu Devin},
  \bibinfo{person}{Sanjay Ghemawat}, \bibinfo{person}{Ian Goodfellow},
  \bibinfo{person}{Andrew Harp}, \bibinfo{person}{Geoffrey Irving},
  \bibinfo{person}{Michael Isard}, \bibinfo{person}{Yangqing Jia},
  \bibinfo{person}{Lukasz Kaiser}, \bibinfo{person}{Manjunath Kudlur},
  \bibinfo{person}{Josh Levenberg}, \bibinfo{person}{Dan Man},
  \bibinfo{person}{Rajat Monga}, \bibinfo{person}{Sherry Moore},
  \bibinfo{person}{Derek Murray}, \bibinfo{person}{Jon Shlens},
  \bibinfo{person}{Benoit Steiner}, \bibinfo{person}{Ilya Sutskever},
  \bibinfo{person}{Paul Tucker}, \bibinfo{person}{Vincent Vanhoucke},
  \bibinfo{person}{Vijay Vasudevan}, \bibinfo{person}{Oriol Vinyals},
  \bibinfo{person}{Pete Warden}, \bibinfo{person}{Martin Wicke},
  \bibinfo{person}{Yuan Yu}, {and} \bibinfo{person}{Xiaoqiang Zheng}.}
  \bibinfo{year}{2015}\natexlab{}.
\newblock \showarticletitle{{TensorFlow: Large-Scale Machine Learning on
  Heterogeneous Distributed Systems}}.
\newblock  \bibinfo{volume}{{1}, 212} (\bibinfo{year}{2015}),
  \bibinfo{pages}{19}.
\newblock
\showURL{%
\url{http://download.tensorflow.org/paper/whitepaper2015.pdf}}


\bibitem[\protect\citeauthoryear{Baeza-yates, Jiang, and Harrison}{Baeza-yates
  et~al\mbox{.}}{2015}]%
        {Baeza-yates2015}
\bibfield{author}{\bibinfo{person}{Ricardo Baeza-yates}, \bibinfo{person}{Di
  Jiang}, {and} \bibinfo{person}{Beverly Harrison}.}
  \bibinfo{year}{2015}\natexlab{}.
\newblock \showarticletitle{{Predicting The Next App That You Are Going To
  Use}}.
\newblock \bibinfo{journal}{{\em Proceedings of the 8th ACM International
  Conference on Web Search and Data Mining\/}} (\bibinfo{year}{2015}),
  \bibinfo{pages}{285--294}.
\newblock
\showISBNx{9781450333177}
\showDOI{%
\url{http://dx.doi.org/10.1145/2684822.2685302}}


\bibitem[\protect\citeauthoryear{Barkan and Koenigstein}{Barkan and
  Koenigstein}{2016}]%
        {Barkan2016}
\bibfield{author}{\bibinfo{person}{Oren Barkan} {and} \bibinfo{person}{Noam
  Koenigstein}.} \bibinfo{year}{2016}\natexlab{}.
\newblock \showarticletitle{{Item2Vec : Neural Item Embedding for Collaborative
  Filtering}}.
\newblock \bibinfo{journal}{{\em Arxiv\/}} (\bibinfo{year}{2016}),
  \bibinfo{pages}{1--8}.
\newblock
\showISSN{16130073}
\showDOI{%
\url{http://dx.doi.org/1603.04259}}


\bibitem[\protect\citeauthoryear{Bemmaor and Glady}{Bemmaor and Glady}{2012}]%
        {Bemmaor2012}
\bibfield{author}{\bibinfo{person}{Albert~C. Bemmaor} {and}
  \bibinfo{person}{Nicolas Glady}.} \bibinfo{year}{2012}\natexlab{}.
\newblock \showarticletitle{{Modeling Purchasing Behavior with Sudden "Death":
  A Flexible Customer Lifetime Model}}.
\newblock \bibinfo{journal}{{\em Management Science\/}} \bibinfo{volume}{{58},
  5} (\bibinfo{date}{5} \bibinfo{year}{2012}), \bibinfo{pages}{1012--1021}.
\newblock
\showISBNx{00251909}
\showISSN{0025-1909}
\showDOI{%
\url{http://dx.doi.org/10.1287/mnsc.1110.1461}}


\bibitem[\protect\citeauthoryear{Cheng, Koc, Harmsen, Shaked, Chandra, Aradhye,
  Anderson, Corrado, Chai, Ispir, Anil, Haque, Hong, Jain, Liu, and Shah}{Cheng
  et~al\mbox{.}}{2016}]%
        {Cheng2016a}
\bibfield{author}{\bibinfo{person}{Heng-Tze Cheng}, \bibinfo{person}{Levent
  Koc}, \bibinfo{person}{Jeremiah Harmsen}, \bibinfo{person}{Tal Shaked},
  \bibinfo{person}{Tushar Chandra}, \bibinfo{person}{Hrishi Aradhye},
  \bibinfo{person}{Glen Anderson}, \bibinfo{person}{Greg Corrado},
  \bibinfo{person}{Wei Chai}, \bibinfo{person}{Mustafa Ispir},
  \bibinfo{person}{Rohan Anil}, \bibinfo{person}{Zakaria Haque},
  \bibinfo{person}{Lichan Hong}, \bibinfo{person}{Vihan Jain},
  \bibinfo{person}{Xiaobing Liu}, {and} \bibinfo{person}{Hemal Shah}.}
  \bibinfo{year}{2016}\natexlab{}.
\newblock \showarticletitle{{Wide {\&} Deep Learning for Recommender Systems}}.
\newblock \bibinfo{journal}{{\em arXiv preprint\/}} (\bibinfo{year}{2016}),
  \bibinfo{pages}{1--4}.
\newblock
\showISBNx{9781450347952}
\showDOI{%
\url{http://dx.doi.org/10.1145/2988450.2988454}}


\bibitem[\protect\citeauthoryear{Covington, Adams, and Sargin}{Covington
  et~al\mbox{.}}{2016}]%
        {Covington2016}
\bibfield{author}{\bibinfo{person}{Paul Covington}, \bibinfo{person}{Jay
  Adams}, {and} \bibinfo{person}{Emre Sargin}.}
  \bibinfo{year}{2016}\natexlab{}.
\newblock \showarticletitle{Deep Neural Networks for YouTube Recommendations}.
  In \bibinfo{booktitle}{{\em Proceedings of the 10th ACM Conference on
  Recommender Systems}} \bibinfo{series}{{\em (RecSys '16)}}. ACM, New York,
  NY, USA, \bibinfo{pages}{191--198}.
\newblock
\showISBNx{978-1-4503-4035-9}
\showDOI{%
\url{http://dx.doi.org/10.1145/2959100.2959190}}


\bibitem[\protect\citeauthoryear{Fader, Hardie, and Lee}{Fader
  et~al\mbox{.}}{2005a}]%
        {Fader2005b}
\bibfield{author}{\bibinfo{person}{Peter~S. Fader}, \bibinfo{person}{Bruce
  G.~S. Hardie}, {and} \bibinfo{person}{Ka~Lok Lee}.}
  \bibinfo{year}{2005}\natexlab{a}.
\newblock \showarticletitle{{Counting Your Customers? the Easy Way: An
  Alternative to the Pareto/NBD Model}}.
\newblock \bibinfo{journal}{{\em Marketing Science\/}} \bibinfo{volume}{{24},
  2} (\bibinfo{year}{2005}), \bibinfo{pages}{275--284}.
\newblock
\showISBNx{07322399}
\showISSN{0732-2399}
\showDOI{%
\url{http://dx.doi.org/10.1287/mksc.1040.0098}}


\bibitem[\protect\citeauthoryear{Fader, Hardie, and Lee}{Fader
  et~al\mbox{.}}{2005b}]%
        {Fader2005a}
\bibfield{author}{\bibinfo{person}{Peter~S. Fader}, \bibinfo{person}{Bruce
  G.~S. Hardie}, {and} \bibinfo{person}{Ka~Lok Lee}.}
  \bibinfo{year}{2005}\natexlab{b}.
\newblock \showarticletitle{{RFM and CLV: Using Iso-Value Curves for Customer
  Base Analysis}}.
\newblock \bibinfo{journal}{{\em Journal of Marketing Research\/}}
  \bibinfo{volume}{{XLII}, November} (\bibinfo{year}{2005}),
  \bibinfo{pages}{415--430}.
\newblock
\showISBNx{00222437}
\showISSN{0022-2437}
\showDOI{%
\url{http://dx.doi.org/10.1509/jmkr.2005.42.4.415}}


\bibitem[\protect\citeauthoryear{Friedman, Hastie, and Tibshirani}{Friedman
  et~al\mbox{.}}{2001}]%
        {hastie2001}
\bibfield{author}{\bibinfo{person}{Jerome Friedman}, \bibinfo{person}{Trevor
  Hastie}, {and} \bibinfo{person}{Robert Tibshirani}.}
  \bibinfo{year}{2001}\natexlab{}.
\newblock \bibinfo{booktitle}{{\em The Elements of Statistical Learning}}.
  \bibinfo{volume}{Vol.~1}.
\newblock Springer Series in Statistics, Springer, Berlin.
\newblock


\bibitem[\protect\citeauthoryear{Grbovic, Radosavljevic, Djuric, Bhamidipati,
  Savla, Bhagwan, and Sharp}{Grbovic et~al\mbox{.}}{2015}]%
        {Grbovic2015}
\bibfield{author}{\bibinfo{person}{Mihajlo Grbovic}, \bibinfo{person}{Vladan
  Radosavljevic}, \bibinfo{person}{Nemanja Djuric}, \bibinfo{person}{Narayan
  Bhamidipati}, \bibinfo{person}{Jaikit Savla}, \bibinfo{person}{Varun
  Bhagwan}, {and} \bibinfo{person}{Doug Sharp}.}
  \bibinfo{year}{2015}\natexlab{}.
\newblock \showarticletitle{{E-commerce in Your Inbox : Product Recommendations
  at Scale Categories and Subject Descriptors}}.
\newblock \bibinfo{journal}{{\em Proceedings of the 21st ACM SIGKDD
  International Conference on Knowledge Discovery and Data Mining\/}}
  (\bibinfo{year}{2015}), \bibinfo{pages}{1809--1818}.
\newblock
\showISBNx{9781450336642}
\showDOI{%
\url{http://dx.doi.org/10.1145/2783258.2788627}}


\bibitem[\protect\citeauthoryear{Hardie, Lin, Kumar, Gupta, Ravishanker,
  Sriram, Hanssens, and Kahn}{Hardie et~al\mbox{.}}{2006}]%
        {Hardie2006}
\bibfield{author}{\bibinfo{person}{B. Hardie}, \bibinfo{person}{N. Lin},
  \bibinfo{person}{V. Kumar}, \bibinfo{person}{S. Gupta}, \bibinfo{person}{N.
  Ravishanker}, \bibinfo{person}{S. Sriram}, \bibinfo{person}{D. Hanssens},
  {and} \bibinfo{person}{W. Kahn}.} \bibinfo{year}{2006}\natexlab{}.
\newblock \showarticletitle{{Modeling Customer Lifetime Value}}.
\newblock \bibinfo{journal}{{\em Journal of Service Research\/}}
  \bibinfo{volume}{{9}, 2} (\bibinfo{year}{2006}), \bibinfo{pages}{139--155}.
\newblock
\showISBNx{10946705}
\showISSN{1094-6705}
\showDOI{%
\url{http://dx.doi.org/10.1177/1094670506293810}}


\bibitem[\protect\citeauthoryear{Jia, Shelhamer, Donahue, Karayev, Long,
  Girshick, Guadarrama, and Darrell}{Jia et~al\mbox{.}}{2014}]%
        {jia2014caffe}
\bibfield{author}{\bibinfo{person}{Yangqing Jia}, \bibinfo{person}{Evan
  Shelhamer}, \bibinfo{person}{Jeff Donahue}, \bibinfo{person}{Sergey Karayev},
  \bibinfo{person}{Jonathan Long}, \bibinfo{person}{Ross Girshick},
  \bibinfo{person}{Sergio Guadarrama}, {and} \bibinfo{person}{Trevor Darrell}.}
  \bibinfo{year}{2014}\natexlab{}.
\newblock \showarticletitle{Caffe: Convolutional Architecture for Fast Feature
  Embedding}. In \bibinfo{booktitle}{{\em Proceedings of the 22nd ACM
  International Conference on Multimedia}}. ACM, \bibinfo{pages}{675--678}.
\newblock


\bibitem[\protect\citeauthoryear{LeCun, Bengio, and Hinton}{LeCun
  et~al\mbox{.}}{2015}]%
        {lecun2015deep}
\bibfield{author}{\bibinfo{person}{Yann LeCun}, \bibinfo{person}{Yoshua
  Bengio}, {and} \bibinfo{person}{Geoffrey Hinton}.}
  \bibinfo{year}{2015}\natexlab{}.
\newblock \showarticletitle{Deep Learning}.
\newblock \bibinfo{journal}{{\em Nature\/}} \bibinfo{volume}{{521}, 7553}
  (\bibinfo{year}{2015}), \bibinfo{pages}{436--444}.
\newblock


\bibitem[\protect\citeauthoryear{McMahan, Holt, Sculley, Young, Ebner, Grady,
  Nie, Phillips, Davydov, Golovin, Chikkerur, Liu, Wattenberg, Hrafnkelsson,
  Boulos, and Kubica}{McMahan et~al\mbox{.}}{2013}]%
        {McMahan2013}
\bibfield{author}{\bibinfo{person}{H~Brendan McMahan}, \bibinfo{person}{Gary
  Holt}, \bibinfo{person}{D Sculley}, \bibinfo{person}{Michael Young},
  \bibinfo{person}{Dietmar Ebner}, \bibinfo{person}{Julian Grady},
  \bibinfo{person}{Lan Nie}, \bibinfo{person}{Todd Phillips},
  \bibinfo{person}{Eugene Davydov}, \bibinfo{person}{Daniel Golovin},
  \bibinfo{person}{Sharat Chikkerur}, \bibinfo{person}{Dan Liu},
  \bibinfo{person}{Martin Wattenberg}, \bibinfo{person}{Arnar~Mar
  Hrafnkelsson}, \bibinfo{person}{Tom Boulos}, {and} \bibinfo{person}{Jeremy
  Kubica}.} \bibinfo{year}{2013}\natexlab{}.
\newblock \showarticletitle{{Ad Click Prediction: A View from the Trenches}}.
\newblock \bibinfo{journal}{{\em Proceedings of the 19th ACM SIGKDD
  International Conference on Knowledge Discovery and Data Mining\/}}
  (\bibinfo{year}{2013}), \bibinfo{pages}{1222--1230}.
\newblock
\showISBNx{9781450321747}
\showISSN{9781450321747}
\showDOI{%
\url{http://dx.doi.org/10.1145/2487575.2488200}}


\bibitem[\protect\citeauthoryear{Mikolov, Chen, Corrado, and Dean}{Mikolov
  et~al\mbox{.}}{2013}]%
        {Mikolov2013}
\bibfield{author}{\bibinfo{person}{Tomas Mikolov}, \bibinfo{person}{Kai Chen},
  \bibinfo{person}{Greg Corrado}, {and} \bibinfo{person}{Jeffrey Dean}.}
  \bibinfo{year}{2013}\natexlab{}.
\newblock \showarticletitle{{Distributed Representations of Words and Phrases
  and their Compositionality}}.
\newblock \bibinfo{journal}{{\em Advances in Neural Information Processing
  Systems\/}} (\bibinfo{year}{2013}), \bibinfo{pages}{3111--3119}.
\newblock
\showISBNx{2150-8097}
\showISSN{10495258}
\showDOI{%
\url{http://dx.doi.org/10.1162/jmlr.2003.3.4-5.951}}


\bibitem[\protect\citeauthoryear{Morrison, Schmittlein, Journal, Statistical,
  and Series}{Morrison et~al\mbox{.}}{1988}]%
        {Journal2016}
\bibfield{author}{\bibinfo{person}{Donald~G Morrison}, \bibinfo{person}{David~C
  Schmittlein}, \bibinfo{person}{Source Journal}, \bibinfo{person}{Royal
  Statistical}, {and} \bibinfo{person}{Society Series}.}
  \bibinfo{year}{1988}\natexlab{}.
\newblock \showarticletitle{{Generalizing the NBD model for Customer Purchases:
  What are the Implications and is it Worth the Effort?}}
\newblock \bibinfo{journal}{{\em Journal of Business {\&} Economic
  Statistics\/}} \bibinfo{volume}{{6}, 1} (\bibinfo{year}{1988}),
  \bibinfo{pages}{129--145}.
\newblock
\showISBNx{07350015}
\showISSN{15372707}
\showDOI{%
\url{http://dx.doi.org/10.1080/07350015.1988.10509648}}


\bibitem[\protect\citeauthoryear{Perozzi and Skiena}{Perozzi and
  Skiena}{2014}]%
        {Perozzi2014}
\bibfield{author}{\bibinfo{person}{Bryan Perozzi} {and} \bibinfo{person}{Steven
  Skiena}.} \bibinfo{year}{2014}\natexlab{}.
\newblock \showarticletitle{{DeepWalk : Online Learning of Social
  Representations}}.
\newblock \bibinfo{journal}{{\em Proceedings of the 20th ACM SIGKDD
  International Conference on Knowledge Discovery and Data Mining\/}}
  (\bibinfo{year}{2014}), \bibinfo{pages}{701--710}.
\newblock
\showISBNx{9781450329569}
\showISSN{9781450329569}
\showDOI{%
\url{http://dx.doi.org/10.1145/2623330.2623732}}


\bibitem[\protect\citeauthoryear{Raiko, Valpola, and LeCun}{Raiko
  et~al\mbox{.}}{2012}]%
        {raiko2012}
\bibfield{author}{\bibinfo{person}{Tapani Raiko}, \bibinfo{person}{Harri
  Valpola}, {and} \bibinfo{person}{Yann LeCun}.}
  \bibinfo{year}{2012}\natexlab{}.
\newblock \showarticletitle{Deep Learning Made Easier by Linear Transformations
  in Perceptrons}. In \bibinfo{booktitle}{{\em Proceedings of the 15th
  International Conference on Artificial Intelligence and Statistics}}. JMLR,
  \bibinfo{pages}{924--932}.
\newblock


\bibitem[\protect\citeauthoryear{Roweis and Saul}{Roweis and Saul}{2000}]%
        {roweis2000nonlinear}
\bibfield{author}{\bibinfo{person}{Sam~T Roweis} {and}
  \bibinfo{person}{Lawrence~K Saul}.} \bibinfo{year}{2000}\natexlab{}.
\newblock \showarticletitle{Nonlinear Dimensionality Reduction by Locally
  Linear Embedding}.
\newblock \bibinfo{journal}{{\em science\/}} \bibinfo{volume}{{290}, 5500}
  (\bibinfo{year}{2000}), \bibinfo{pages}{2323--2326}.
\newblock


\bibitem[\protect\citeauthoryear{Schmittlein, Morrison, and
  Colombo}{Schmittlein et~al\mbox{.}}{1987}]%
        {Schmittlein1987}
\bibfield{author}{\bibinfo{person}{David~C. Schmittlein},
  \bibinfo{person}{Donald~G. Morrison}, {and} \bibinfo{person}{Richard
  Colombo}.} \bibinfo{year}{1987}\natexlab{}.
\newblock \showarticletitle{{Counting Your Customers: Who Are They and What
  Will They Do Next?}}
\newblock \bibinfo{journal}{{\em Management Science\/}} \bibinfo{volume}{{33},
  1} (\bibinfo{year}{1987}), \bibinfo{pages}{1--24}.
\newblock
\showISBNx{00251909}
\showISSN{0025-1909}
\showDOI{%
\url{http://dx.doi.org/10.1287/mnsc.33.1.1}}


\bibitem[\protect\citeauthoryear{Vanderveld, Pandey, Han, and
  Parekh}{Vanderveld et~al\mbox{.}}{2016}]%
        {Vanderveld2016}
\bibfield{author}{\bibinfo{person}{Ali Vanderveld}, \bibinfo{person}{Addhyan
  Pandey}, \bibinfo{person}{Angela Han}, {and} \bibinfo{person}{Rajesh
  Parekh}.} \bibinfo{year}{2016}\natexlab{}.
\newblock \showarticletitle{{An Engagement-Based Customer Lifetime Value System
  for E-commerce}}.
\newblock \bibinfo{journal}{{\em Proceedings of the 22nd ACM SIGKDD
  International Conference on Knowledge Discovery and Data Mining\/}}
  (\bibinfo{year}{2016}).
\newblock
\showISBNx{9781450342322}


\bibitem[\protect\citeauthoryear{Vasile, Smirnova, and Conneau}{Vasile
  et~al\mbox{.}}{2016}]%
        {Vasile2016a}
\bibfield{author}{\bibinfo{person}{Flavian Vasile}, \bibinfo{person}{Elena
  Smirnova}, {and} \bibinfo{person}{Alexis Conneau}.}
  \bibinfo{year}{2016}\natexlab{}.
\newblock \showarticletitle{{Meta-Prod2Vec - Product Embeddings Using
  Side-Information for Recommendation}}. In \bibinfo{booktitle}{{\em
  Proceedings of the 10th ACM Conference on Recommender Systems - RecSys '16}}.
  \bibinfo{pages}{225--232}.
\newblock
\showISBNx{9781450340359}
\showDOI{%
\url{http://dx.doi.org/10.1145/2959100.2959160}}


\bibitem[\protect\citeauthoryear{Vincent, Larochelle, Bengio, and
  Manzagol}{Vincent et~al\mbox{.}}{2008}]%
        {vincent2008extracting}
\bibfield{author}{\bibinfo{person}{Pascal Vincent}, \bibinfo{person}{Hugo
  Larochelle}, \bibinfo{person}{Yoshua Bengio}, {and}
  \bibinfo{person}{Pierre-Antoine Manzagol}.} \bibinfo{year}{2008}\natexlab{}.
\newblock \showarticletitle{Extracting and Composing Robust Features with
  Denoising Autoencoders}. In \bibinfo{booktitle}{{\em Proceedings of the 25th
  International Conference on Machine Learning}}. ACM,
  \bibinfo{pages}{1096--1103}.
\newblock


\bibitem[\protect\citeauthoryear{Wangperawong, Brun, and
  Pavasuthipaisit}{Wangperawong et~al\mbox{.}}{2016}]%
        {Wangperawonga}
\bibfield{author}{\bibinfo{person}{Artit Wangperawong},
  \bibinfo{person}{Cyrille Brun}, {and} \bibinfo{person}{Rujikorn
  Pavasuthipaisit}.} \bibinfo{year}{2016}\natexlab{}.
\newblock \showarticletitle{{Churn Analysis Using Deep Convolutional Neural
  Networks and Autoencoders}}.
\newblock \bibinfo{journal}{{\em arXiv preprint arXiv\/}}
  (\bibinfo{year}{2016}), \bibinfo{pages}{1--6}.
\newblock


\bibitem[\protect\citeauthoryear{Zadrozny and Elkan}{Zadrozny and
  Elkan}{2001}]%
        {zadrozny2001obtaining}
\bibfield{author}{\bibinfo{person}{Bianca Zadrozny} {and}
  \bibinfo{person}{Charles Elkan}.} \bibinfo{year}{2001}\natexlab{}.
\newblock \showarticletitle{Obtaining Calibrated Probability Estimates from
  Decision Trees and Naive Bayesian Classifiers}. \bibinfo{journal}{{\em
  Proceedings of the 18th International Conference on Machine Learning\/}}
  \bibinfo{volume}{1} (\bibinfo{year}{2001}), \bibinfo{pages}{609--616}.
\newblock


\end{thebibliography}

\end{document}